\documentclass[logo,copyright,12pt]{googledeepmind}

\newcommand{\systemlong}{\textrm{AlphaProof Nexus}}
\newcommand{\AP}{\textrm{AlphaProof}}
\newcommand{\erdos}{{Erd\H{o}s}}
\newcommand{\AEv}{\textrm{AlphaEvolve}}

\newcommand{\multiset}[1]{\{\!\{#1\}\!\}}
\newcommand{\card}[1]{\left|#1\right|}
\newcommand{\degG}[2]{\deg_{#1}(#2)}

\makeatletter
\renewcommand{\paragraph}{%
  \@startsection{paragraph}{4}%
  {\z@}{0.8ex plus 0.2ex minus 0.1ex}{-1em}%
  {\normalfont\normalsize\bfseries}%
}
\makeatother

\title{Advancing Mathematics Research with AI-Driven Formal Proof Search}

\author[1$^\dagger$]{George Tsoukalas}
\author[1$^\dagger$]{Anton Kovsharov}
\author[1$^\dagger$]{Sergey Shirobokov}
\author[1]{Anja Surina}
\author[1]{Moritz Firsching}
\author[2]{Gergely B\'erczi}
\author[1]{Francisco J. R. Ruiz}
\author[1]{Arun Suggala}
\author[1]{Adam Zsolt Wagner}
\author[1]{Eric Wieser}
\author[1]{Lei Yu}
\author[1]{Aja Huang}
\author[1]{Mikl\'os Z. Horv\'ath}
\author[1]{Andrew Ferraiuolo}
\author[1]{Henryk Michalewski}
\author[1]{Edward Lockhart}
\author[3]{Codrut Grosu}
\author[1]{Thomas Hubert}
\author[1]{Matej Balog}
\author[1]{Pushmeet Kohli}
\author[1$^\dagger$]{Swarat Chaudhuri}
\affil[1]{Google DeepMind\footnote{$^\dagger$Equal contributions. The first three authors are in random order. Correspondence to \url{pushmeet@google.com} and \url{swarat@google.com}.}}
\affil[2]{Aarhus University}
\affil[3]{Google}

\begin{abstract} \bfseries \boldmath
Large language models (LLMs) increasingly excel at mathematical reasoning, but their unreliability limits their utility in mathematics research. A mitigation is using LLMs to generate formal proofs in languages like Lean. We perform the first large-scale evaluation of this method's ability to solve open problems. Our most capable agent autonomously resolved 9 of 353 open {\erdos} problems at the per-problem cost of a few hundred dollars, proved 44/492 OEIS conjectures, and is being deployed in combinatorics, optimization, graph theory, algebraic geometry, and quantum optics research. A basic agent alternating LLM-based generation with Lean-based verification replicated the {\erdos} successes but proved costlier on the hardest problems. These findings demonstrate the power of AI-aided formal proof search and shed light on the agent designs that enable it.
\end{abstract}

\begin{document}
\maketitle

\section{Introduction}

Large language models (LLMs) have recently shown remarkable promise in solving complex mathematics problems \cite{feng2026autonomous, woodruff2026accelerating}, but \emph{unreliability} remains a primary barrier to their integration into mathematics research. Because LLM-generated natural language proofs can contain subtle logical errors, or ``hallucinations,'' they require expensive expert review. Mistakes in unreviewed intermediate steps can cascade through a proof, limiting the complexity of tasks that can be delegated to AI. 

Recent efforts \cite{hubert2025olympiad, achim2025aristotle} mitigate these issues by using AI to generate proofs in formal languages like Lean \cite{moura2021lean}, in which a compiler automatically verifies every logical step. So far, successes of this paradigm have been concentrated in competition mathematics and the human-aided formalization of natural language arguments \cite{hariharan2026milestone}. 
In this paper, we demonstrate its broader potential through a large-scale evaluation on open research-level problems. 

To this end, we developed a framework, {\systemlong}, for LLM-aided proof generation and used it to build a basic agent in which a set of subagents independently searches for proofs with feedback from the Lean compiler. We also developed a ``full-featured'' agent in which subagents are coordinated using an evolutionary algorithm \cite{novikov2025alphaevolve} and can use {\AP}~\cite{hubert2025olympiad}, a system for olympiad-level Lean theorem-proving based on reinforcement learning, as a focused proof tool.

Our full-featured agent autonomously solved 9 {\erdos} problems out of 353 attempted, including two questions that had been open for 56 years \cite{SCHOEN2001191, Baier2004ANO, elsholtz2016erd}, at the inference cost of a few hundred dollars per problem. It also 
proved 44/492 open conjectures from the Online Encyclopedia of Integer Sequences (OEIS), resolved an open question on Hilbert functions in algebraic geometry, improved an open bound in convex optimization by discovering a novel algorithmic parameter schedule, identified several misformalizations in the literature, helped resolve an open problem from Ben Green's well-known list \cite{green2024100}, and is aiding ongoing research efforts on quantum optics and graph theory. 

To understand the impact of the agent design on these results, we did a post-hoc analysis of the performance of the full-featured and basic agents, as well as two agents with intermediate capabilities, on the 9 {\erdos} problems solved by the full-featured agent. Remarkably, the basic agent solved all 9 problems, though at a higher cost on the harder problems. 

Overall, our results demonstrate the utility of LLM-powered formal proof search as a tool for mathematics research, and point to an ongoing shift from specialized trained systems toward simple agentic loops as LLMs become more capable. All Lean proofs and select natural-language proofs are available in \url{https://www.github.com/google-deepmind/alphaproof-nexus-results}.

\begin{figure}[htbp]
      \centering
\includegraphics[width=0.9\textwidth]{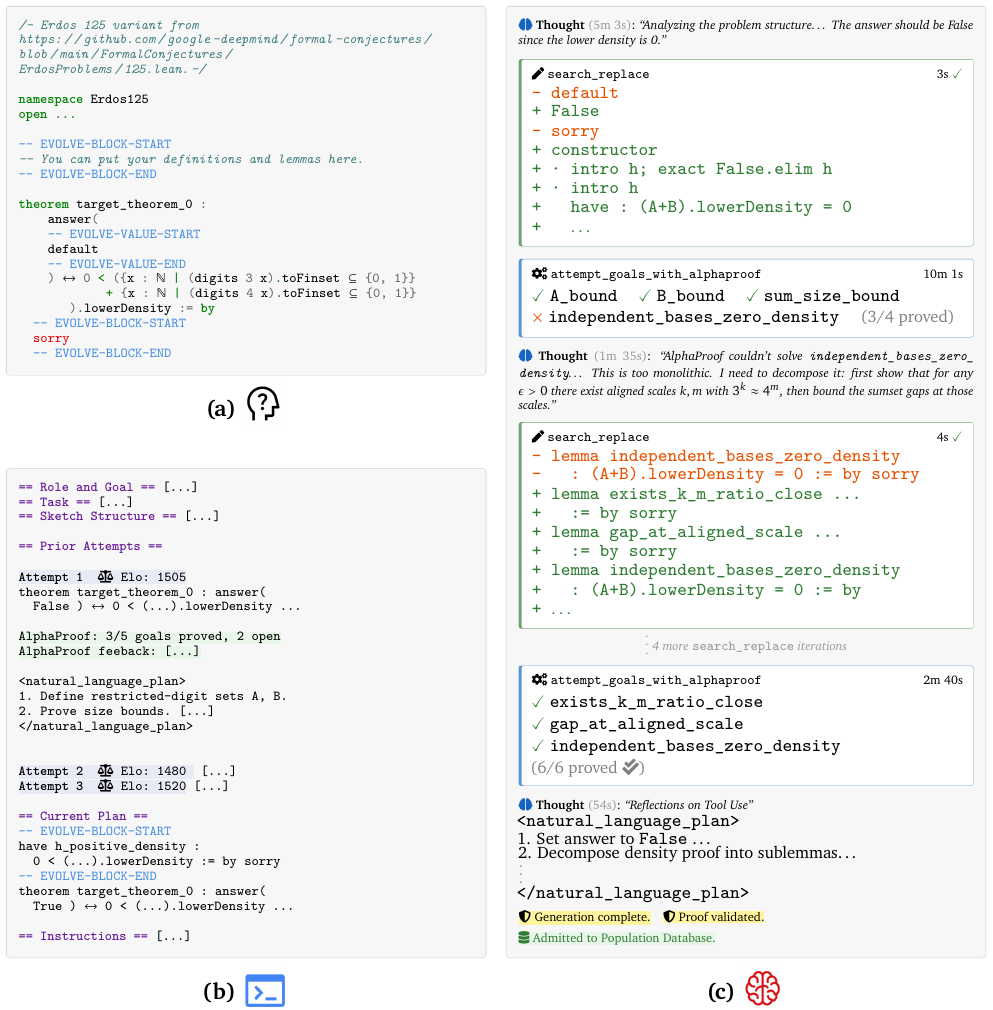} 
\caption{\textbf{Example inputs/outputs for an {\AP}-equipped agent (applied to {\erdos} \#125).}  The user provides a Lean file with a specification of the problem, and an empty proof body replaced with the \texttt{sorry} placeholder. \textbf{(a)} Modifications are permitted only within \texttt{EVOLVE-BLOCK} and \texttt{EVOLVE-VALUE} markers. 
\textbf{(b)} During sketch refinement, the prover subagent is shown an assembled prompt template with the current proof, and optionally prior attempts/sketches, their Elo ratings, and feedback from {\AP}'s attempts on unsolved goals. \textbf{(c)} The prover reasons about the problem informally and invokes 
tools.
In this example, the prover invoked {\AP} which resolved all but one goal. 
The prover then decomposed that goal into three simpler lemmas, and called {\AP} again, which then resolved all remaining goals. The agent also produced a natural language summary of its attempt at the end of generation. 
}
    \label{fig:components}
\end{figure}

\section{{\systemlong}}

\paragraph{Lean.} Lean \cite{moura2021lean} is a proof assistant in which definitions, theorems, and proofs are all mechanically verified code. Proofs are constructed via a sequence of applications of \emph{tactics}, or elementary proof steps. Lean's compiler ``executes'' a proof tactic-by-tactic, tracking the proof goals pending after every tactic. A proof is correct if it leads the compiler to a state with no pending goals.

Lean includes a special $\mathtt{sorry}$ tactic, which immediately closes pending goals while passing the type checker. Proving a theorem thus amounts to the generation of type-safe code without $\mathtt{sorry}$ tactics. 

\paragraph{Input and Output.} {\systemlong} is a new framework for agents that query frontier LLMs and the Lean compiler. The agents take as input a Lean file consisting of a target theorem with $\mathtt{sorry}$ in place of a proof, along with definitions and library imports on which the theorem depends (we refer to such a file as a \emph{proof sketch}). Optionally, the user may input additional natural-language context and domain knowledge encoded in Lean. 

The input sketch is annotated with user-provided markers that delineate which code segments the agent may modify (see Fig.~\ref{fig:components}).
Within \texttt{EVOLVE-BLOCK} markers, the agent can introduce helper lemmas, definitions, and proof steps. \texttt{EVOLVE-VALUE} markers are used to enclose expressions (e.g., parameters) whose values the agent can change. On successful termination, the agent outputs a $\mathtt{sorry}$-free proof of the target theorem.

\paragraph{Agent Architecture.} 
The basic agent in {\systemlong} (agent (A)) consists of a set of \emph{prover subagents} that execute independently with no shared state. Each subagent 
is a ``Ralph loop''~\cite{huntley2025ralph} consisting of a sequence of \emph{episodes}: multi-turn LLM inference loops (based on Gemini 3.1 Pro) in which the subagent can reason via chain-of-thought and refine the sketch using a search-and-replace tool. After each turn, the subagent uses Lean to check that the current proof sketch compiles; if it does not, Lean's error message is used to direct the next turn. If the sketch contains \texttt{sorry} when the episode terminates, the subagent adds a comment to the sketch summarizing the lessons learned from the episode. The resulting sketch is the input to the next episode. 

We extended the basic agent into one (B) that can query {\AP} \cite{hubert2025olympiad} to fill out missing parts of sketches. Queries to {\AP} can return a proof, disproof (a proof that the submitted subgoal is false), or a failure message; proofs are directly substituted into the sketch, while disproofs and failure messages are fed into the prover's prompt.

\begin{figure}[htbp]
    \centering
    \includegraphics[width=0.8\textwidth]{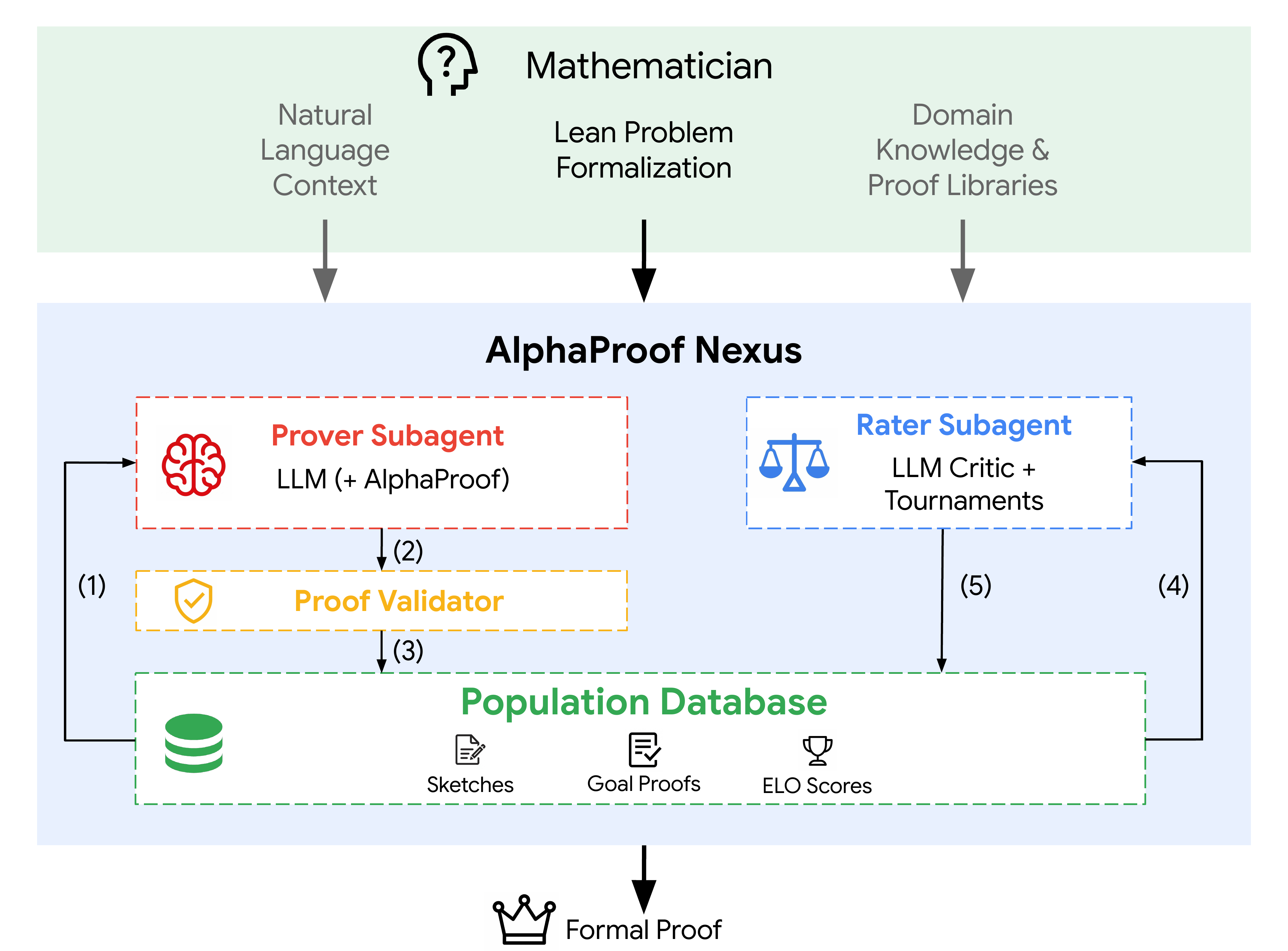} 
    \caption{\textbf{Design of the full-featured {\systemlong} agent.} The mathematician provides as input a Lean theorem with \texttt{sorry} for a proof, and optionally, natural language context and additional domain knowledge encoded in Lean. The agent architecture consists of a basic generation-validation pipeline and an optional evolutionary framework. An LLM-based prover subagent attempts to solve the problem by refining proof sketches \textbf{(2)}. The subagent may optionally call {\AP} as a tool; each invocation of {\AP} on a goal returns whether a proof or disproof was found, or whether it was unsuccessful in resolving the goal. The sketch produced in the end is checked by a validator to ensure the problem statement was not changed unsafely and the proof compiles. If all goals are successfully proved, the agent outputs the final Lean proof. This basic pipeline can be extended with an evolutionary population database and rating mechanism. In this configuration, validated sketches are admitted into the Population Database \textbf{(3)}. Simultaneously, rater subagents sample previous attempts \textbf{(4)}, which are then ranked by an LLM critic in matches. The match results are admitted back into the database \textbf{(5)} to update the Elo scores of the sketches. These scores are then used by the evolutionary algorithm to sample prior sketches from the database, constructing an overall prompt -- which includes any optional inputs from the mathematician -- to condition new episodes \textbf{(1)}.}
    \label{fig:detailed_system}
\end{figure}

Separately, we developed an evolutionary agent (C), inspired by AlphaEvolve \cite{novikov2025alphaevolve}, in which prover subagents sample from and contribute to a shared population database of sketches. A challenge here is the mismatch between evolutionary algorithms, which typically assume a graduated fitness landscape, and formal proof evaluation, which is inherently binary. To bridge this gap, we used a pool of \emph{rating agents} (based on the less expensive Gemini 3.0 Flash) to construct relative rankings of sketches based on their plausibility, clarity, and novelty. We aggregated these rankings into Elo ratings for the sketches and used a sampling procedure based on the P-UCB formula \cite{hubert2021learning,silver2017masteringchessshogiselfplay} to drive the search. 

Finally, we combined the {\AP} and evolution capabilities into a ``full-featured'' agent (D; see Fig.~\ref{fig:detailed_system}) in which prover subagents can use {\AP} as well as evolutionary search. We used this agent as the instrument in our exploration of open research problems. 

\section{Systematic Evaluation on Open Problems}

\paragraph{{\erdos} Problems.} 

Bloom maintains an online catalog \cite{erdosproblems} of over 1200 open problems posed by Paul {\erdos} and his collaborators.
The open-source Formal Conjectures repository \cite{deepmind2026formalconjectures} contains Lean formalizations of a subset of these problems.
We ran Agent (D) on all these formal statements (353 at the time of the run), terminating the search if no proof was found within 3000 episodes. The agent solved 9/353 problems (Table~\ref{tab:major_results} and supplementary text); after each solve, experts on our team validated that the Lean statement faithfully captured the original conjecture. We have publicly shared the proofs, and our results have been logged on Terence Tao's wiki on AI contributions to {\erdos} problems \cite{TaoErdosWiki}. 

Several proofs require sophisticated constructions and the synthesis of distinct mathematical arguments. For example, problem \#12(i), posed by {\erdos} and S\'ark\"ozy in 1970 \cite{Erds1970OnTD}, asks if there is an infinite set $A$ that satisfies a restrictive divisibility constraint -- no element may divide the sum of two larger elements -- and which satisfies the density condition $\liminf_{N \to \infty} \frac{|A \cap [1, N]|}{\sqrt{N}} > 0$; this problem had received attention in multiple prior works \cite{elsholtz2016erd, SCHOEN2001191, Baier2004ANO}. Our agent constructed $A$ as an infinite union of disjoint ``blocks'' $B_i \subseteq [P_i, 1.1P_i]$ for a suitably rapidly growing sequence $(P_i)$. To satisfy the local divisibility conditions, the proof integrates the Chinese Remainder Theorem with the properties of sets that avoid length-3 arithmetic progressions. 

Problem \#125 concerns the sumset $A+B$, where $A$ is the set of non-negative integers whose base-3 representation uses only the digits 0 and 1, and $B$ is the analogous set in base 4.  
The question of whether the lower density of $(A + B)$ is positive was open since 1996 \cite{BurrErdos125}. 
Our agent resolved the conjecture by synthesizing an inductive thinning argument that exploits the Diophantine proximity of the two multiplicatively independent bases ($3^m \approx 4^k$). 


The agent also served as a tool for detecting and fixing misformalizations. For example, in {\erdos} problems \#125 and \#741(i), the interpretation of ``density'' in the original informal statements was amended to ``lower density'' and ``upper density'' respectively, after our full-featured agent found proofs using density as ``natural density.'' Following the correction of the ambiguity, the agent was still able to resolve the questions.

\noindent \textit{Failure Analysis.} We analyzed the highest-scoring sketches (measured by Elo) across a random sample of problems on which our agent failed. First, the agent frequently offloaded a problem's core difficulty into a single \texttt{sorry} within a helper lemma that reiterated the target statement in a slightly different form. Explicitly prompting against this behavior failed to prevent it. Second, for several problems, the top sketches relied on lemmas marked with \texttt{sorry} that the agent claimed were established results in the mathematical literature. Upon manual inspection, these lemmas proved to be hallucinations. These failure modes underscore the value of end-to-end formal verification.

\paragraph{OEIS.} We also applied the agent to systematically explore open problems in the OEIS \cite{oeis}, a massive repository of integer sequences and their known and open properties. We used Gemini to autoformalize 492 open questions from the OEIS and applied the agent to the resulting Lean statements. As a guard against misformalization, the agent was required to prove ``test lemmas'' verifying the first few terms of each sequence against its formal definition before attempting the target conjectures. The agent found proofs for 44 conjectures that a manual review found to be correctly formalized and previously unproven. Two of the proofs appear in the supplementary material.

\begin{table}[t]
\centering
\footnotesize
\renewcommand{\arraystretch}{1.4}
\setlength{\tabcolsep}{8pt}
\begin{tabular}{l p{6.5cm} p{6.5cm}}
\hline
\textbf{ID} & \textbf{Conjecture Summary} & \textbf{Proof Technique} \\ \hline
12\,(i) & (1970) Existence of $A \subset \mathbb{Z}^+$ s.t.\ $a \nmid (b+c)$ with $|A \cap [1,N]|/\sqrt{N} > \delta > 0$ infinitely often & Block-based construction via CRT and 3-AP avoiding sets \\
12\,(ii) & (1970) Existence of $A \subset \mathbb{Z}^+$ s.t.\ $a \nmid (b+c)$ with $|A \cap [1,N]| \gg N^{1-\epsilon}$ for any $\epsilon$ and $N$ sufficiently large & Block-based construction via CRT and 3-AP avoiding sets \\
125 & (1996) $\{ \sum \epsilon_k 3^k | \epsilon_k \in \{0,1\}\} + \{ \sum \epsilon_j 4^j | \epsilon_j \in \{0,1\}\}$ has lower density zero & Inductive thinning via Diophantine approx. $3^m \approx 4^k$  \\
138$^*$ & (1981) Van der Waerden numbers satisfy $W(k{+}1) - W(k) \to \infty$ & Greedy coloring extension with monochromatic intersection lemma \\
152 & (1994) Sufficiently large Sidon sets $A$ contain many isolated points in $A+A$ & Elementary argument by bounds on interior points, shifted neighbors and quadruples \\
741\,(i) & (1994) If $A + A$ has upper density, there exists a decomposition $A = A_1 \sqcup A_2$ s.t. $A_1 + A_1, A_2 + A_2$ have positive upper density & Bounding of upper density of $A_i + A_i$ via cases on upper density of $A$ \\
741\,(ii) & (1994) Existence of a basis $A$ of order $2$ with $A_1 + A_1$, $A_2 + A_2$ having bounded gaps for all $A_1, A_2$ such that $A = A_1 \sqcup A_2$ &  Explicit construction via rapidly growing sequence \\
846 & (1992) Existence of infinite set $A$ such that any finite subset contains many non-collinear points, but is not a finite union of non-collinear sets & Label $K_\infty$ vertices with terms from fast-growing sequence, apply map on edges $\{i,j\} \mapsto (x_i + x_j, x_i^2 + x_ix_j + x_j^2)$ to obtain $A$ \\
26$^{*, \dagger}$ & (1995) Existence of $A$ such that upper density of $A+k$ is $< 1 - \frac{1}{4}$ for all naturals $k$ & Block-based construction using increasing sequence of primes, CRT, and bounding u. density by cases \\
\hline
\end{tabular}
\caption{\textbf{Open problems from the ErdosProblems repository autonomously resolved by our full-featured agent.} The asterisk indicates a variant of the main problem with this number. Problem \#26 (annotated with $\dagger$) is a more general variant of a question posed by {\erdos}, but was not posed by {\erdos} himself.}
\label{tab:major_results}
\end{table}

\section{Deployment in Mathematics Research}

\paragraph{Optimization Theory.} Agent (D) resolved an open question in optimization: proving an exact $\mathcal{O}(1/t)$ convergence rate for the Anchored Gradient Descent-Ascent (GDA) algorithm for min-max convex-concave optimization, thereby tightening the slower bound established by \cite{ryu2019ode}. 
The proof departs from continuous-time ordinary differential equations (ODE) analysis used in previous work, instead using a discrete-time recurrence-based approach. The agent did not merely verify a fixed algorithm: we marked the learning schedule as a parameter within an \texttt{EVOLVE-VALUE} block in the input file, allowing the agent to simultaneously search for the schedule and the proof, ultimately discovering a novel parameter choice that yields the stronger guarantee. We previously released a deformalized version of the proof as a preprint \cite{surina2026improvedlastiterateconvergencerate}. 
Subsequent work \cite{cai2026last} has extended this result.

\paragraph{Graph Theory.} The \emph{graph reconstruction conjecture}
\cite{graphreconstruction} is one of the oldest open problems in combinatorics.
It asserts that every finite simple graph with at least three vertices is
determined, up to isomorphism, by the multiset of its vertex-deleted subgraphs,
known as its \emph{deck}.  AlphaEvolve \cite{novikov2025alphaevolve} experiments helped us formulate two bipartite variants of the graph reconstruction conjecture, along with a proposed full reconstruction algorithm. 
Agent (D) proved the correctness of the proposed
algorithm under an additional type condition on the bipartite graph.  The
corresponding unrestricted bipartite reconstruction statement remains open; in
that case, the agent
generated proof sketches and
strategies that helped clarify the structure of the problem and led to
simplified reformulations of the conjectures originally suggested by
AlphaEvolve.

Separately, the agent proved an open graph theory conjecture regarding a bound on the maximum number of leaves over all spanning trees of a graph $G$, relating it to the maximum number, over all vertices $v \in G$, of independent sets in the neighborhood of $v$. The problem was posed by Graffiti \cite{Graffiti}, an automated conjecturing system, in 1996, and points to an interesting future opportunity to close the loop between AI-based conjecturing and proof.

\paragraph{Algebraic Geometry.} 

We evaluated Agent (D) on eight algebraic geometry problems, ranging from textbook-style exercises to open research questions on Hilbert functions, and it solved two of the four open problems. One of these problems was Zanello’s conjecture on log-concavity of pure O-sequences \cite{Zanello2024}. Pure O-sequences arise as Hilbert functions of monomial Artinian level algebras, and their study goes back to work from the 1970s on Hilbert functions of graded algebras \cite{Stanley1978}. Zanello’s conjecture belongs to a broader program on shape and positivity properties of pure O-sequences. For sequences with codimension 3 and type 2, the weaker unimodality statement had previously been proved \cite{BMMNZ2012}. The agent's proof establishes log-concavity in this case, which is surprising because Zanello had shown that log-concavity fails for many other families of pure O-sequences. Moreover, in the broader class of level Hilbert functions, where the defining algebra need not be monomial and hence the Hilbert function need not be a pure O-sequence, Zanello showed that non-log-concave examples exist in all remaining codimension-type pairs except the still-open case (3,2). The agent-generated argument is substantial, using a nontrivial reformulation of the Hilbert function and a detailed case analysis of the resulting second-difference inequalities.

We also deployed Agent (D) within a larger effort on the AI-driven formalization of the Stacks Project algebraic geometry textbook \cite{stacksproject}. A key challenge here is bridging the abstraction gap between the textbook's natural-language definitions, theorems, and proofs and their formal Lean counterparts. 
Once the necessary definitions and theorem statements had been formalized, we used Agent (D) to automatically construct several intricate proofs. The agent was especially useful in cases where the source material gave only a brief outline or where the material is not widely published, such as \href{https://stacks.math.columbia.edu/tag/04KM}{Lemma 04KM} on separable field extensions.

\paragraph{Additive Combinatorics.} Our agent helped resolve problem \#57 from Green's well-known list of open conjectures \cite{green2024100}. The problem asks whether two specific quadratically structured function spaces coincide. The functions of interest map elements of an Abelian group $G$ to the complex numbers. Here, our agent autonomously solved a variant of the problem in which the functions are real-valued, but a personal communication from Green clarified that the complex-valued case was the intended challenge. While the agent could not immediately prove the intended version of the problem, numerical heuristics using floating-point arithmetic provided a candidate counterexample (the cyclic group $\mathbb{Z}/3\mathbb{Z}$ and a specific separating functional). We formalized the problem of whether this counterexample disproves the correct conjecture, and the agent autonomously proved that it indeed does. A paper on the result is in the works \cite{firsching2026strict}. 

\paragraph{Quantum Optics.} With Mario Krenn, we investigated a set of quantum optics problems concerning the existence of monochromatic quantum graphs with $N$ vertices and $d$ colors drawn from domains such as the reals, the complex numbers, and $\{-1, 0, 1\}$. These constructions correspond to $N$-particle quantum states with local Hilbert space dimension $d$ -- in particular, high-dimensional Greenberger–Horne–Zeilinger (GHZ) states realizable via linear optics \cite{Krenn_2017}. Our agent resolved multiple conjectures of this form, in particular with $N = d \in \{4,6,10\}$. A paper on these results is in preparation \cite{Krenn2026TensorAlgebraic}.

\begin{figure}[htbp]
    \centering
    \includegraphics[width=1\textwidth]{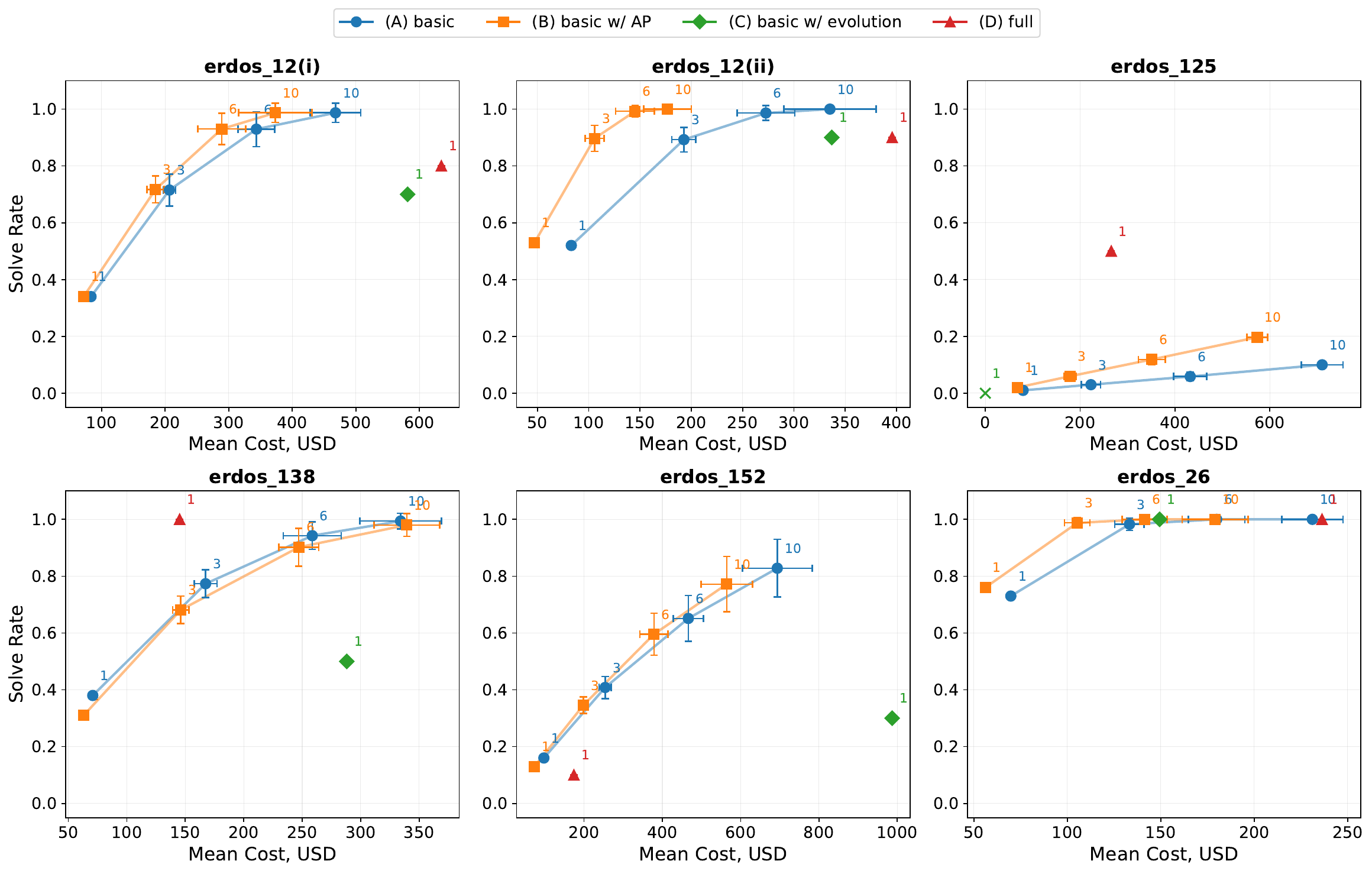} 
    \caption{\textbf{Solve rate versus mean inference cost (USD) across six {\erdos} problem instances.} The solve rates are evaluated for the four agents: (A) basic (blue circles), (B) basic with {\AP} (orange squares), (C) basic with evolution (green diamonds), and (D) full-featured (red triangles). Numeric annotations denote the number of independent attempts $K \in \{1,3,6,10\}$ grouped together; error bars indicate one standard error interval. Due to their higher costs, agents (C) and (D) lack variance estimates, as independent attempts were used to obtain a single point estimate (see main text for details). For agents (A) and (B), each curve traces the cost–performance Pareto frontier as $K$ increases, revealing diminishing marginal returns at higher budgets. Agent (B) generally matches or exceeds the solve rate of the basic configuration at comparable cost. Costs reported for (B) and (D) do not include the inference cost of {\AP}. While for most problems configuration (A) or (B) is the best, for some challenging problems like {\erdos} \#125, the full-featured configuration (D) performs significantly better. Note that accounting for the estimated {\AP} cost of 60 USD does not change the above outcomes beyond the margin of error.}
    \label{fig:scaling_plot_erdos_main}
\end{figure}

\section{Impact of Agent Architecture and Model}

Exploration of a large space of open problems is expensive, and we chose the full-featured agent (agent (D)) for this based on its strong performance on competition benchmarks. To understand which architectural components are necessary for its successes, we compared its performance against agents (A), (B), and (C) on the {\erdos} problem set in Table~\ref{tab:major_results}. 

We compared the agents by analyzing the solve rate against the cost (in US dollars) per successfully proven problem. 
We report computational cost in USD because it directly measures the barrier to reproducing this work and provides a natural common currency for comparing agents that allocate compute differently (e.g., agent (D) uses Gemini 3.0 Flash for rater subagents and Gemini 3.1 Pro for provers). We do not intend USD as a comparison of AI and human mathematical labor.

For agents (C) and (D), in which a single proof attempt requires 10 subagents, 
we executed 10 attempts per problem. In contrast, for the basic Agent (A) and its {\AP}-equipped extension (B), we ran 100 independent attempts, each with a single subagent. 

Because (A) and (B) consist of independent subagents, we simulated scenarios where they have $K$ subagents by grouping the attempts into chunks of size $100/K$; a chunk was considered successful if any attempt included in the chunk proved the statement. The overall solve rate was defined to be the fraction of successful chunks. To calculate the monetary cost for these successful chunks, we identified the earliest timestamp $T$ of a successful attempt and summed the costs of all attempts within that chunk up to $T$. Since 
agents (C) and (D) are more expensive, we used the independent attempts to obtain a single point estimate.

Fig.~\ref{fig:scaling_plot_erdos_main} compares the agents across six {\erdos} problems (results for the rest are in the supplementary material). 
Agents (A) and (B) perform similarly -- within the margin of error -- on four of the problems, though agent (B) is more efficient on problems 12(ii) and 125. Agent (D) outperforms (A) and (B) on problems 138 and 125, offering significant monetary savings (2x to 5x), but is roughly half as cost-efficient on the remaining problems. 
We also compared agents (A), (B), and (D) on the wall-clock time needed to solve the problems. The results broadly followed the inference cost trends, with (B) offering savings over (A) in several of the problems, and (D) substantially outperforming both (A) and (B) on problems 138 and 125.

Next, we evaluated {\AP} in standalone tree-search mode and versions of agent (A) based on smaller models (Gemini 3.0 Flash, Gemini 3.1 Flash-Lite). These systems could not solve any of the problems. Finally, we evaluated two commercial coding agents -- Codex (GPT-5.5) and Claude Code (Opus 4.7) -- in ``goal'' mode on the problems. Codex solved 7/9 problems, failing to solve Problems \#152 and \#125 within 12 hours. We could not get Claude Code to solve any of the problems. 

\paragraph{Cost and Variance.} Per-problem inference costs exhibit high variance due to the stochastic nature of our agents. The reported costs also do not capture the full cost of discovery: we applied the full-featured agent to all 353 {\erdos} problems in Formal Conjectures, and identifying tractable problems was itself a significant computational investment. {\AP} cost approximately 27.5 TPU hours (\$60 USD) per problem on v6e TPUs.

\section{Discussion}

We have provided a large-scale demonstration of the value of formal proof search agents on research-level mathematical problems. 
Recently, some natural-language reasoning systems have been shown to succeed in research-level mathematical tasks, including {\erdos} problems \cite{feng2026aletheia, alexeev2026short2}. However, the use of AI-generated informal proofs, either as standalone products or as inputs to a subsequent formalization stage, requires careful validation by human experts. Formal verification can serve as a filter for determining which proofs merit human review. 

The effectiveness of our basic agent in our post-hoc analysis was surprising. At the time we were planning our large-scale exploration, simpler agentic loops did not show strong performance on competition-level benchmarks, and this informed our decision to use the full-featured agent. The LLM landscape has since shifted substantially. We attribute the basic agent's success to both this shift and the power of compiler feedback in grounding LLM reasoning. The full-featured agent retains an advantage on the hardest problems for now. However, as LLM capabilities grow, this advantage may diminish.

At present, our agents' successes are concentrated in areas such as combinatorics, convex optimization, and number theory, where Lean's mathematics library \cite{mathlib2020} is mature and tasks often decompose into tractable subgoals. Even most {\erdos} problems remain out of reach, let alone problems that require extensive new theory. Additionally, our agents inherit the biases of their underlying LLMs and exhibit high search variance. Characterizing the agents' boundaries and expanding them is an important direction for future work.

We built {\systemlong} with the belief that the future of mathematics lies in human-machine partnership, where interactive AI tools serve to expand a mathematician's creative capacity. Our results support this vision. Our mathematician collaborators found that proof attempts by our agents enhanced their understanding of a problem, even when an agent could not prove the claim at hand. Because the sketches were formal, experts could focus on the unresolved subgoals rather than re-verifying the entire argument. Moreover, the agents were powerful tools for detecting misformalizations. These experiences suggest that AI-driven formal proof search can serve not only to solve problems but to deepen human understanding.


\bibliography{refs} 
\bibliographystyle{plain}


\section*{Acknowledgments}

We thank Emilien Dupont, Dan Roy, and Daniel Zheng for their careful feedback on the paper, Alexander Novikov for help with LLM infrastructure, Katerina Hristova for help with Lean formalization, and Sebastian Nowozin and Taylan Cemgil for their guidance on designing an Elo scoring mechanism for proof sketches.

\paragraph*{Author contributions:}
SC, SS, and GT conceptualized and implemented the first version of {\systemlong}. AK led the engineering and infrastructure decisions. AK, SS, GT, and A.Surina developed the final version of {\systemlong} with inputs from FJR, AF, and SC. AK and SS maintained the underlying infrastructure of \systemlong. 

GT initiated and led the use of the system in mathematics research tasks. AK, SS, GT, A.Surina, LY, and SC coordinated the large-scale runs on open research problems. GT, MF, GB, A.Suggala, AZW, and EL identified problems to solve and validated the generated solutions. GB, in particular, was the first research mathematician to use {\systemlong}. MF, GT, A.Surina, FJR, HM, SC, and EW contributed to Lean formalizations of problems. SS led the systematic analysis of agent design choices, with additional contributions from AK, MZH and AH. 

SC, SS, GT, FJR, and A.Surina wrote the paper with inputs from MF, AZW, HM, and AF. SS, AF, and EW created the figures with inputs from SC, GT, A.Surina and FJR. GB, AZW, MF, GT, A.Surina and CG wrote the natural language proofs, with HM providing agent-generated drafts to facilitate the writing process.  MZH and EW assisted with Lean setup and integration with AlphaProof. AF coordinated the creation of the accompanying Github repository. 

MB and TH provided technical advice throughout the project. SC led the overall effort, with PK providing sponsorship and technical and strategic direction. 

\paragraph*{Competing interests:} There are no competing interests to declare.

\paragraph*{Data and materials availability:} All Lean proofs are available in the accompanying repository 
\url{https://www.github.com/google-deepmind/alphaproof-nexus-results}. Natural language proofs
of some key results are provided in the repository and the supplementary text.
\newpage 



\appendix

\section{Materials and Methods}

\subsection{Further Details on {\systemlong} Agents}

\begin{figure}[htbp]
\centering 
\includegraphics[scale=1.1]{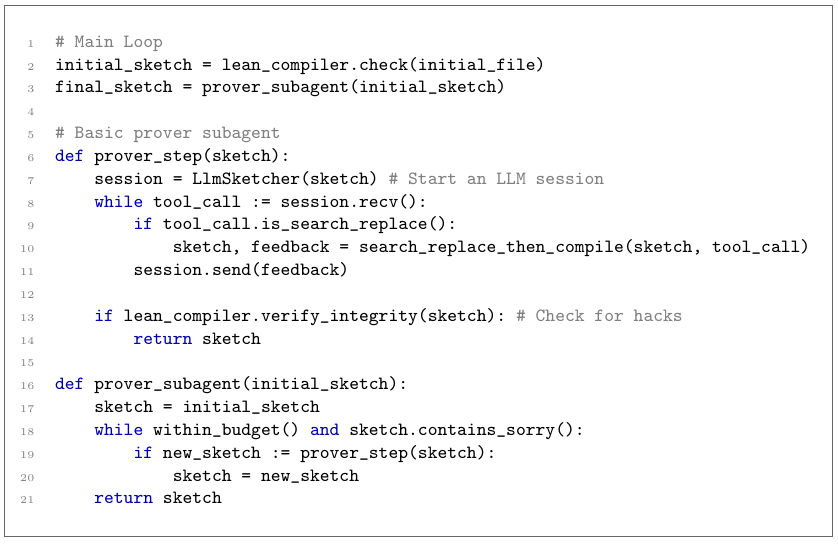}
\caption{\textbf{Pseudocode for the basic agent (A).} A prover subagent executes a sequence of steps in a loop. Each step is a conversation with a LLM instance (Gemini 3.1 Pro). During this conversation, the subagent can view a Lean file and apply search-replace edits. After each change, the Lean compiler provides feedback, such as compilation errors. As for agent (D), we check for the integrity of the final sketch once the conversation is over. $N$ subagents run independently in parallel and stop as soon as one of them found a valid proof.}
\label{fig:basic_pseudocode}
\end{figure}

\paragraph{Basic Agent.} The pseudocode for the basic agent (A) is shown in Figure~\ref{fig:basic_pseudocode}. The agent operates $N$ independent subagents that share no state. All subagents start with the same initial proof sketch. As soon as one subagent finds a proof, all others are terminated.

Each subagent implements a Ralph loop \cite{huntley2025ralph} consisting of a sequence of episodes. The full prompt is shown in Figure~\ref{fig:basic_prompt}. Within each episode, the subagent runs a multi-turn session with an access to a \texttt{search\_replace} tool. After each edit, the Lean code is compiled and compiler feedback is passed back to the model. When the subagent ends an episode, the code is validated using SafeVerify \cite{SafeVerify2025}, which checks the proof against the theorem specification and guards against environment exploits (e.g., axiom injection). If validation succeeds and the proof is \texttt{sorry}-free, the proof is returned. If \texttt{sorry} remains, the subagent summarizes lessons learned in a comment and begins the next episode from the current sketch. If validation fails, the subagent reverts to the previous sketch.

\begin{figure}[htbp]
\centering 
\includegraphics{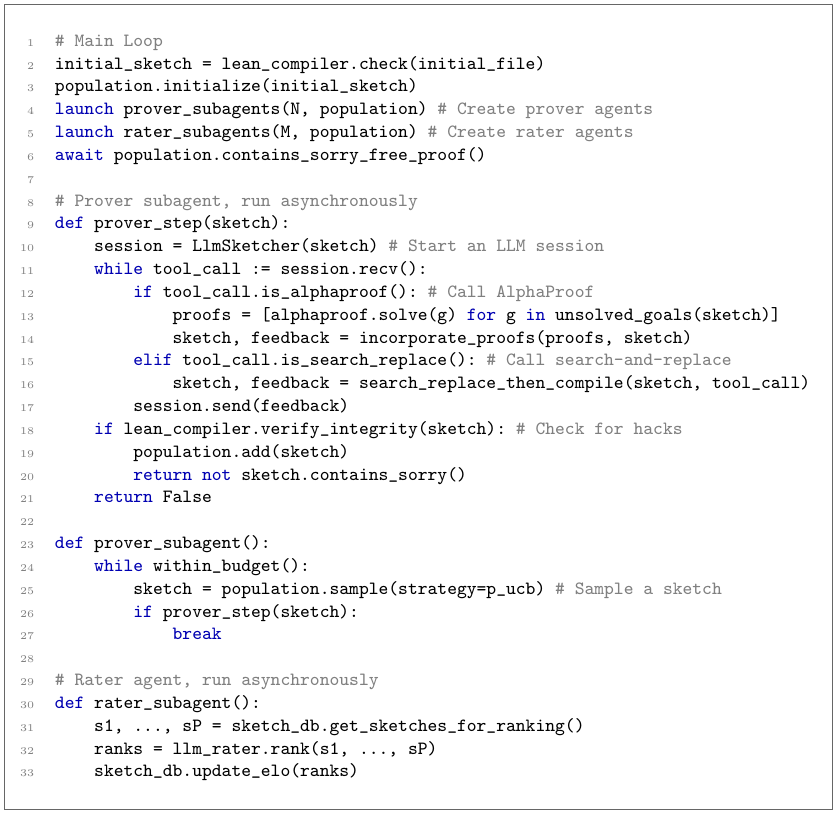}
\caption{\textbf{Pseudocode for the main components of the full-featured agent (Agent (D)).} The main loop creates a pool of asynchronous prover subagents and raters and awaits the creation of a full (\texttt{sorry}-free) proof. At each step, a prover subagent uses the P-UCB strategy to sample a parent sketch from the population and initiates a stateful conversation with a LLM instance (Gemini 3.1 Pro). The agent has access to two tools: {\AP} and a structured search-and-replace operation. The results of these tool calls provide context for the subsequent turn. Once the conversation concludes, the resulting sketch is added to the population only if it compiles successfully and the original theorem remains uncompromised. Concurrently, each \texttt{rater\_agent} samples $P$ sketches, uses a LLM (Gemini 3.0 Flash) to determine which is the most promising, and updates the Elo scores used for P-UCB sampling. We set $P=7$ in all the experiments.}
\label{fig:pseudocode}
\end{figure}

\paragraph{Full-featured Agent.} The pseudocode for the full-featured agent (D) is shown in Figure~\ref{fig:pseudocode}. The evolutionary process is orchestrated by a controller, which executes a continuous, asynchronous loop. At each step, the controller performs the following stages: 

\begin{enumerate} 
\setlength{\itemsep}{8pt}
\setlength{\parsep}{2pt}
\item \textbf{Database sampling:} The controller selects a root proof sketch $S_{\text{root}}$ by sampling from the database, along with $M = 2$ auxiliary inspiration sketches ${S_{\text{insp}}}$. The selection strategy balances exploitation of high-rated sketches with exploration of diverse candidates (see ``Population Database and Matchmaking'').

\item \textbf{Prompt construction:} A prompt $\mathcal{X}$ is assembled to guide the LLM. It integrates the formal problem specification, the Lean source code and natural language plan of $S_{\text{root}}$, and structured feedback derived from \AP's previous attempts on $\{S_{\text{insp}}\}$. As in \AEv, the controller encourages diversity by stochastically injecting instructions such as ``decompose unsolved goals,'' ``combine ideas from prior attempts,'' or ``try a completely new approach.''

\item \textbf{Prover subagent:} The assembled prompt $\mathcal{X}$ is dispatched to the LLM (Gemini 3.1 Pro), initiating a multi-turn episode. To scale to large Lean files, the subagent outputs mutations via a \texttt{search\_replace} tool in a compact diff format rather than rewriting the entire file. The subagent can also query {\AP} to test specific subgoals mid-episode; the feedback indicates whether the goal was proven, disproven, or unresolved. To manage compute, each episode is restricted to a maximum of 5 {\AP} queries and 90 search-and-replace edits. At the conclusion of the episode, the generated sketch undergoes a sandbox check that permits \texttt{sorry} placeholders but verifies that the original target theorem statement was not altered.

\item \textbf{Validation:} Once the candidate sketch $S'$ passes the sandbox check, it undergoes formal validation. The system extracts all remaining \texttt{sorry} subgoals and cross-references them against a global goal cache using a deep hash of their exact Lean state (see ``Global Goal Caching''). If a subgoal was previously resolved, the proof is retrieved immediately; otherwise, it is dispatched to {\AP}. If {\AP} closes all remaining goals, the fully assembled, \texttt{sorry}-free proof is passed to SafeVerify \cite{SafeVerify2025} for final validation, ensuring the proof compiles and that no disallowed axioms (including \texttt{sorryAx}) were introduced. If any subgoals remain unresolved, $S'$ is registered with its remaining \texttt{sorry} placeholders.

\item \textbf{Database registration:} The candidate $S'$, along with per-subgoal feedback from {\AP}, is registered in the database. Its fitness is then determined asynchronously via Elo matchmaking.
\end{enumerate}

{\AP} has a Test-Time Reinforcement Learning (TTRL) mode in which it learns to solve a problem by solving its AI-generated variants at inference time; however, we prioritize the use of compute for LLM inference and run {\AP} in its low-compute tree search inference mode.

\paragraph{Population Database and Matchmaking.} Unlike systems where numerical fitness can be directly computed (e.g., empirical runtime), formal proof evaluation yields only discrete signals: whether the code compiles and whether the proof is complete. Agent (D) overcomes this by decoupling generation from fitness assignment, using LLM-based relative review to evaluate the promise of incomplete sketches.

\paragraph{Elo-based Rating.} Asynchronous rater agents (Gemini 3.0 Flash) continuously sample sets of $P = 7$ sketches from the database for pairwise ``matches.'' We found $P=7$ to provide a good trade-off between information per LLM call and input context size. Each rater produces a relative ranking by evaluating the clarity of the proof strategy, the plausibility of remaining goals, and the mathematical novelty of the approach.

We model match outcomes using a Plackett-Luce distribution \cite{plackett1975analysis,luce1959individual}, in which each sketch $s$ has a latent strength parameter $\lambda_s$. We place a hierarchical prior $p(\lambda_s | r_s) = \textrm{Gamma}(1, r_s)$ with rate parameter $r_s$ distributed as $p(r_s) = \textrm{Gamma}(1,1)$. This distribution has heavier tails than a simple Gamma over $\lambda_s$ while keeping the distributions conditionally conjugate; we expect the choice of hyperparameters to have a relatively small impact. 

We infer sketches' posterior strengths using a Gibbs sampling procedure \cite{caron2012efficient}. For each sketch $s$, we draw $I$ Gibbs samples $\{\lambda_s^{(i)}\}_{i=1}^{I}$ from the posterior distribution of that sketch's strength parameter, obtained after drawing $B$ burn-in samples (we set $I=1000$ and $B=200$ based on an experiment with synthetic data). We then use the posterior sample mean $\lambda_s^{\textrm{mean}} = \frac{1}{I} \sum_i \lambda_s^{(i)}$ to obtain the sketch's Elo score as
\[
    \textrm{Elo}_s = 1200 + 400 \log_{10} \lambda_s^{\textrm{mean}}.
\]

To obtain the set of $P$ sketches for each match, we use a Thompson sampling strategy that repeatedly takes an independent sample $\lambda_s$ for each sketch and chooses the sketch with the highest value. In practice, we obtain the independent samples via Gibbs sampling, retaining only every $25$-th sample in order to mitigate in-chain correlation. This is sampling with replacement, so the same sketch may appear more than once. After sampling exactly $P$ times, we remove duplicate sketches from the chosen set, replacing them with the sketches with the highest posterior sample variance $\lambda_s^{\textrm{var}}$.

Occasionally the LLM raters may output ties between the input sketches. Since the Plackett-Luce model does not consider ties, we break ties randomly by sampling from the model.

\paragraph{Evolutionary Selection.}
During the sampling phase, the controller selects ``parent'' sketches using the Predictor + Upper Confidence Bound (P-UCB) formula. To focus the search and reduce computational overhead, the sampler first filters the population to the top 64 highest-scoring sketches based on their current Elo ratings. For these top candidates, the Elo ratings are normalized to a range of $[0, 1]$ to yield a base score $q$. The final P-UCB score for each sketch is then computed as:
\[
\text{score} = q + c \frac{\sqrt{\sum V_i}}{v + 1}
\]
where $v$ is the number of times the specific sketch has been visited (sampled), $\sum V_i$ is the total number of visits across the filtered population, and $c$ is a tunable exploration constant (set at $0.2$ for this work). This mechanism prioritizes the exploitation of the most promising sketches -- those within the elite top-64 threshold -- while the UCB exploration bonus ensures adequate exploration of newly promoted or infrequently sampled candidates within that elite tier, preventing the search from collapsing into a single, suboptimal lineage. The values $c=0.2$ and top-64 were chosen empirically based on observed performance. 

\paragraph{Global Goal Caching and Incorporation.}
Independent proving agents can generate the same goals to dispatch to {\AP}. For efficiency, Agent (D) implements a global goal cache within the database. When a sketch is parsed, the system computes a deep hash of the exact formal Lean context and target for every generated subgoal (the \texttt{goal\_id}). Before querying {\AP}, the validator checks this cache. If a specific subgoal's state was already proved or formally disproved in any prior sketch across the population, the result -- along with the specific tactic sequence or value function -- is retrieved and incorporated into the current sketch. Novel subgoals are batched and dispatched concurrently to {\AP} via non-blocking remote procedure calls (RPCs), and their subsequent results are cached to accelerate future generations.

\paragraph{{\AP} Budget.}
When a novel subgoal is evaluated, {\AP} executes a formal tree search to discover a proof or disproof. To prevent the system from stalling on intractable or hallucinated goals, {\AP} operates under a strict computational budget, typically restricted to $400$ simulations and bounded by a hard RPC timeout. We compile {\AP}'s response into textual feedback in case it cannot solve a subgoal. This feedback is then associated with this particular subgoal and rendered in the corresponding prompt sketches.

\paragraph{Implementation.}
The entire {\systemlong} infrastructure is implemented in Python, utilizing the \texttt{asyncio} framework. The evolutionary controller in Agent (D) runs an event loop that distributes work across asynchronous threads for generation, validation, and Elo rating. Formal validation and compilation are executed inside isolated sandboxes (Docker instances) running Lean v4.27 and Pantograph \cite{aniva2025pantographmachinetomachineinteractioninterface}. This ensures that generated code is type-checked in a secure, stateful environment without risking execution of malicious code. The LLM backend leverages an ensemble of models: Gemini 3.1 Pro is used for the complex reasoning required in the multi-turn proving agent, while the faster Gemini 3.0 Flash is deployed for high-throughput match rating and evaluation synthesis.

\subsection{Prompts}

\begin{figure}[htbp]
\centering 
\includegraphics[scale=0.9]{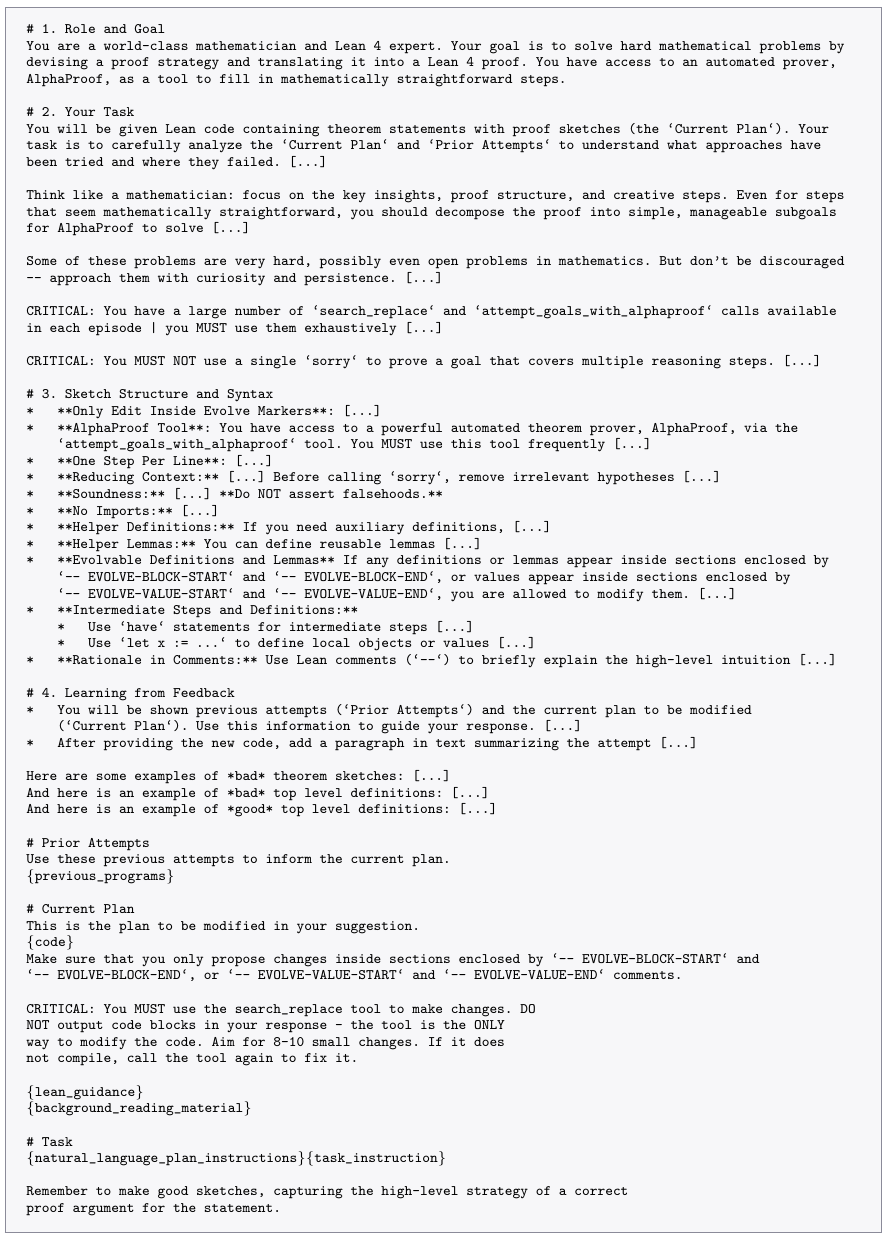}
\caption{\textbf{Prompt for prover subagents in the full-featured agent (D) (condensed).} Elided text is represented by \texttt{[\ldots]}. Text in braces denotes template variables populated at runtime. For example, \texttt{\{code\}} is replaced by the current Lean file.}
\label{fig:sketcher-prompt}
\end{figure}

\begin{figure}[htbp]
\centering
\includegraphics[scale=0.9]{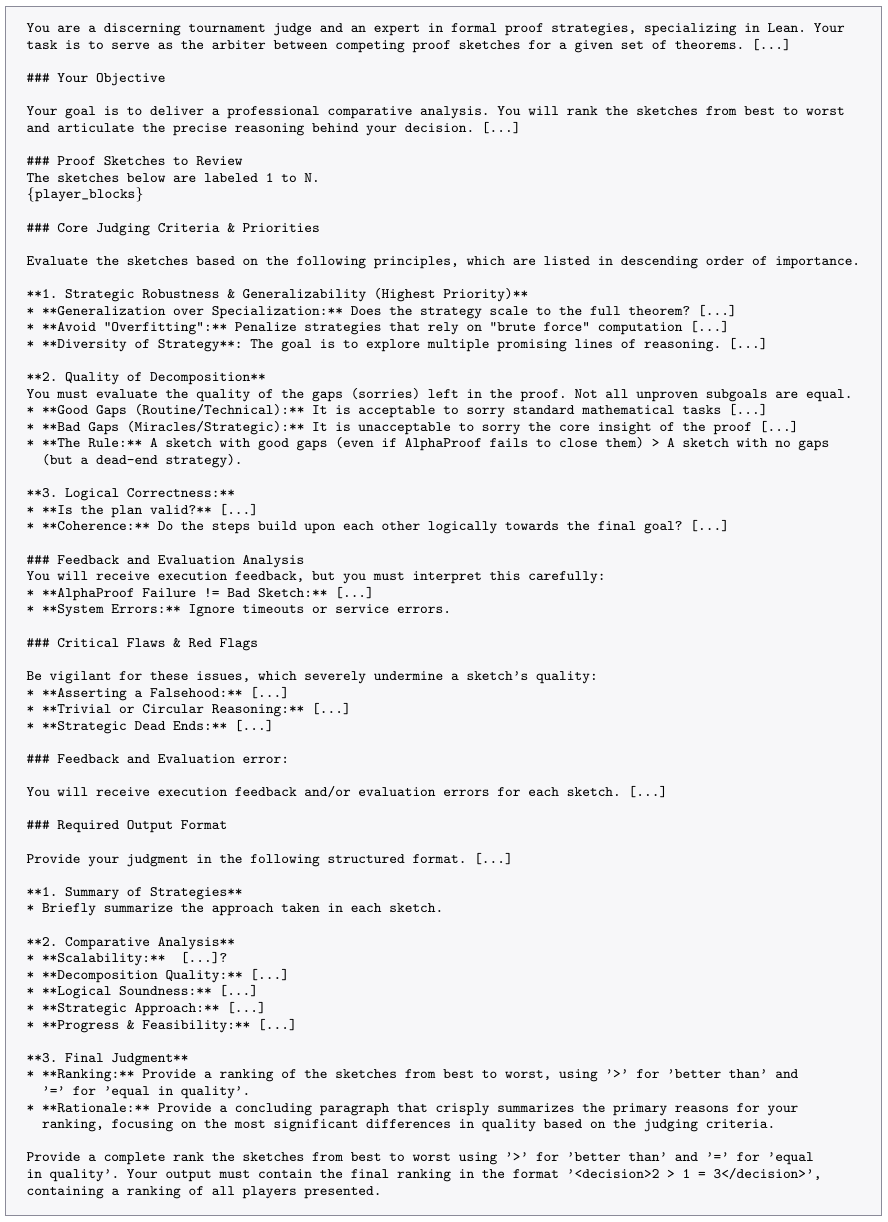}
\caption{\textbf{Prompt for raters in the full-featured agent (D) (condensed).} Elided text is represented by \texttt{[\ldots]}. Text in braces denotes template variables populated at runtime. For example, \texttt{\{player\_blocks\}} is replaced by the sketches to be compared.}
\label{fig:rater-prompt}
\end{figure}

\begin{figure}[htbp]
\centering 
\includegraphics[scale=1.1]{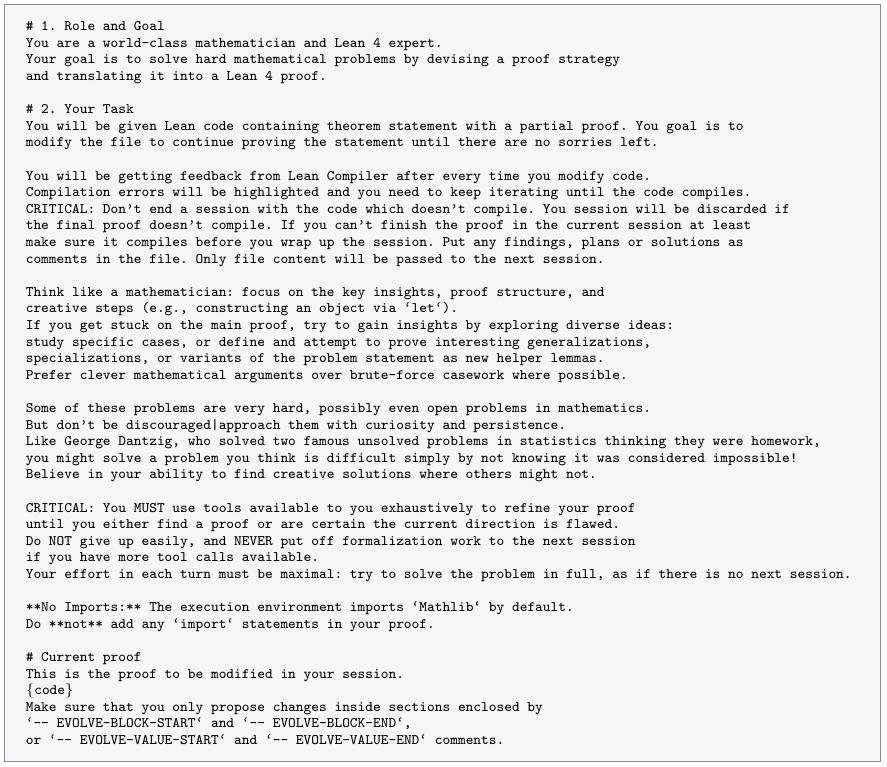}
\caption{\textbf{Full prompt for the basic agent (A).} Text in braces denotes template variables populated at runtime. For example, \texttt{\{code\}} is replaced by the current Lean file.}
\label{fig:basic_prompt}
\end{figure}

Figure~\ref{fig:sketcher-prompt} shows the main pieces of the prompt used for prover subagents in the full-featured agent (D). Figure~\ref{fig:rater-prompt} shows the prompt for the rating subagents in the full-featured agent (D). Figure~\ref{fig:basic_prompt} shows the full prompt for the basic agent (A).


\section{Supplementary Text}

\subsection{Related Work}

\paragraph{Formal Theorem Proving with LLMs.} 
There is a large literature on neural-network-guided search over machine-checkable formal proofs \cite{yang2024formal,chen2025seed,deepseekv2,wang2025kimina,lin2025goedel}. Early work such as GPT-$f$ \cite{polu2020generative}
established the viability of language models in this setting; subsequent systems improved tactic generation, premise selection,
and interaction with external provers \cite{jiang2022thor,yang2023leandojo}. The Draft-Sketch-Prove system
\cite{jiang2022draft} introduced the hierarchical approach of generating informal sketches before translating them into formal
steps, and recent systems such as Hilbert \cite{varambally2025hilbert} and Aristotle
\cite{achim2025aristotle} further separate high-level proof planning from low-level elaboration. {\AP} \cite{hubert2025olympiad} first showed that reinforcement learning (RL) could elevate formal theorem-proving to the Math Olympiad level. Some of the successes of that system came from using test-time RL; however, the system also supports a lower-cost tree search inference mode that we leverage. Subsequently, several other systems have demonstrated strong performance in competition mathematics \cite{achim2025aristotle,chen2025seed,lin2025goedel,axle2025}.

At the research level, AI-aided formal proofs have been used primarily to verify results derived in natural language (either by human mathematicians or AI systems), rather than to discover new ones. In particular, Aristotle \cite{achim2025aristotle} was used to formalize AI-generated natural language proofs of several {\erdos} problems -- see Tao's wiki \cite{TaoErdosWiki} for more details. Gauss \cite{mathinc2026gauss} 
was used to produce a formalization of Viazovska's proof of sphere packing in dimension 8 \cite{hariharan2026milestone}. Our work differs in that we use Lean as a medium of novel mathematical discovery.

\paragraph{Evolution for Mathematical Discovery.}
The use of evolutionary algorithms to search over programs has a long history in AI \cite{koza1994genetic}.  FunSearch \cite{Romera-Paredes2024} introduced the idea of using LLM-guided evolution to search for mathematical constructions represented as code. FunSearch was later extended into AlphaEvolve \cite{novikov2025alphaevolve}, which has been used to improve bounds and
find novel constructions in numerous distinct areas of mathematics \cite{georgiev2025mathematical,nagda2026reinforced, nagda2025reinforced}.
Our implementation of agent (D) reuses several components of AlphaEvolve. However, the fundamental difference between the two systems is that AlphaEvolve aims to find programs that optimize a quantitative reward function, while the goal of our agents is to find proofs that pass a boolean formal verification criterion.

\paragraph{Natural Language Proof Discovery.} 
A large body of recent work explores whether LLMs can perform research-level mathematical tasks in natural language, as evidenced in the AI Co-Mathematician \cite{zheng2026aicomathematician}. Focusing on theorem proving, Aletheia \cite{feng2026autonomous} exemplifies this approach via heavy test-time compute with interleaved
generation and revision. Other provers include FullProof \cite{bryan2026motivic} and DeepThink \cite{deepmind2026deepthink}. A proprietary model developed by OpenAI resolved several {\erdos} conjectures 
\cite{alexeev2026short,alexeev2026short2}. Subsequently, {\erdos} problem \#1196 was resolved informally through community experimentation using GPT-5.4 \cite{alexeev2026primitivesetsvonmangoldt}.
Many proofs and partial results discovered this way have been subsequently autoformalized with the help of agents such as Aristotle \cite{achim2025aristotle} and Gauss \cite{mathinc2026gauss}. Further collaborations between mathematicians and AI models have yielded
results in optimization theory \cite{jang2025point}, algebraic geometry \cite{schmitt2025extremal,bryan2026motivic}, and spectral
theory \cite{jain2026equality}.

\subsection{Selection of {\erdos} and OEIS Problems}

For the {\erdos} evaluation, we ran our agent on all Lean statements of {\erdos} problems available in the Formal Conjectures repository \cite{deepmind2026formalconjectures} as of early February 2026 -- 353 problems in total. We did not select which problems to attempt; the set was determined entirely by what the open-source community had formalized from the 1200+ problems catalogued on the ErdosProblems site. We recognize that this process has a bias toward problems amenable to formalization in Lean.

For the OEIS evaluation, we began with a corpus of 2649 open conjectures drawn from the OEIS \cite{oeis}. We prompted Gemini to select 500 problems that are 
non-trivial, mathematically interesting, not famous open problems, and good candidates for automated theorem-proving, and used a Gemini-based agent to formalize them. Of the 500 problems, 8 were excluded due to incompatibilities introduced by a Lean version upgrade, yielding a final set of 492 problems.

\subsection{Details on Comparisons across Agent Architectures and Models}
\begin{figure}[htbp]
    \centering
    \includegraphics[width=1\textwidth]{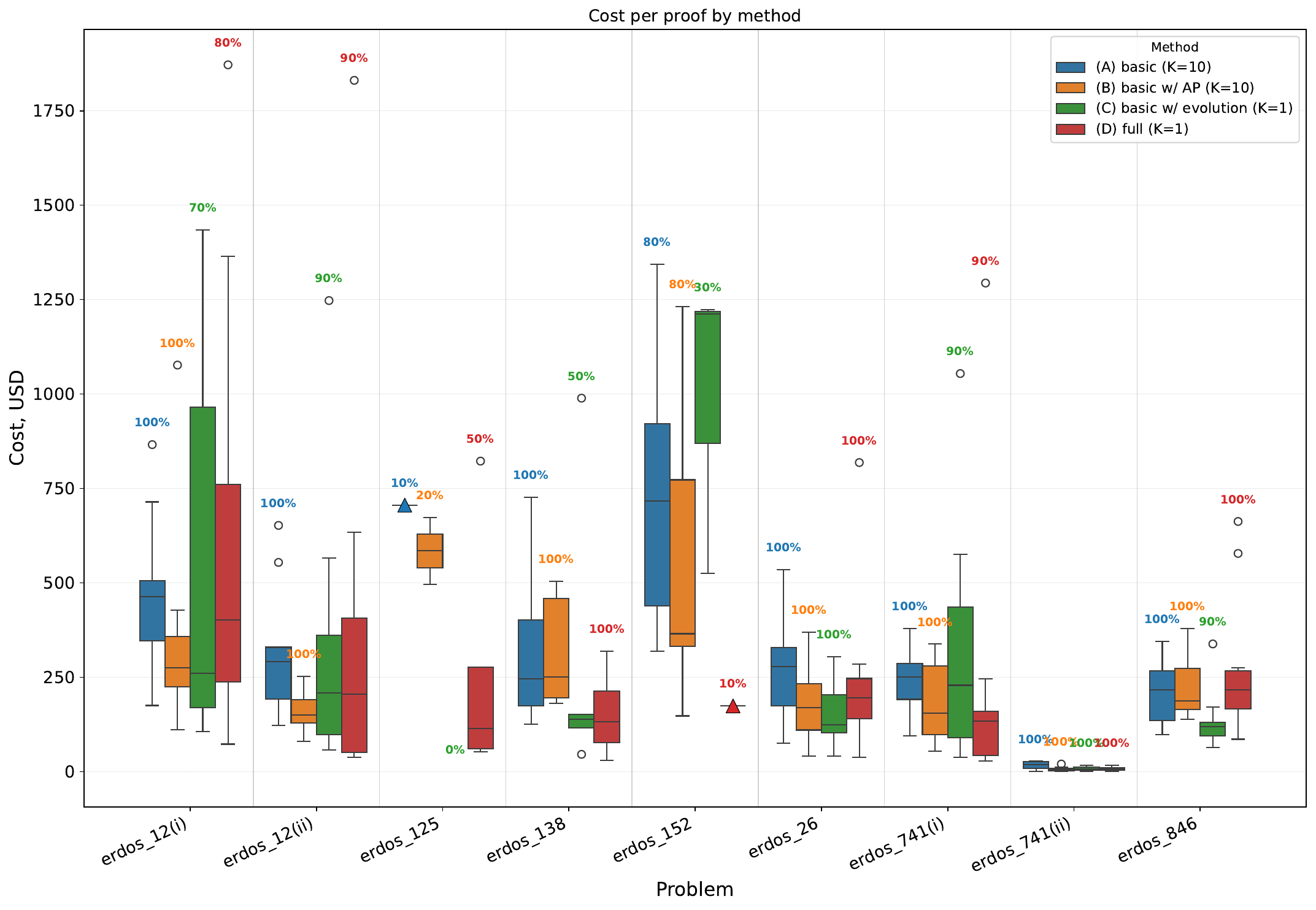} 
    \caption{\textbf{Box plots illustrating the distribution of a cost for successful proof for the {\erdos} problems.} Four distinct experimental configurations are compared: basic@K=10 (blue), basic with {\AP} @K=10 (orange), basic with evolution (green), and full (red). The solid horizontal line within each box denotes the median cost, while the upper and lower box boundaries represent the third and first quartiles, respectively. Whiskers extend to the rest of the distribution, and individual circles indicate outlier data points. Triangular markers denote instances with a single data point where only one proof attempt was successful.}
    \label{fig:cost_distribution_appendix}
\end{figure}

Figure~\ref{fig:cost_distribution_appendix} reports, for each {\erdos} problem, the distribution of computational costs for successful runs, along with the solve rate for each agent. We note a large variance in the cost for most problems and methods, which highlights the stochastic nature of the agent. This is especially noticeable on problems such as {\erdos} 12(ii) and 152.

\begin{figure}[htbp]
    \centering
    \includegraphics[width=1\textwidth]{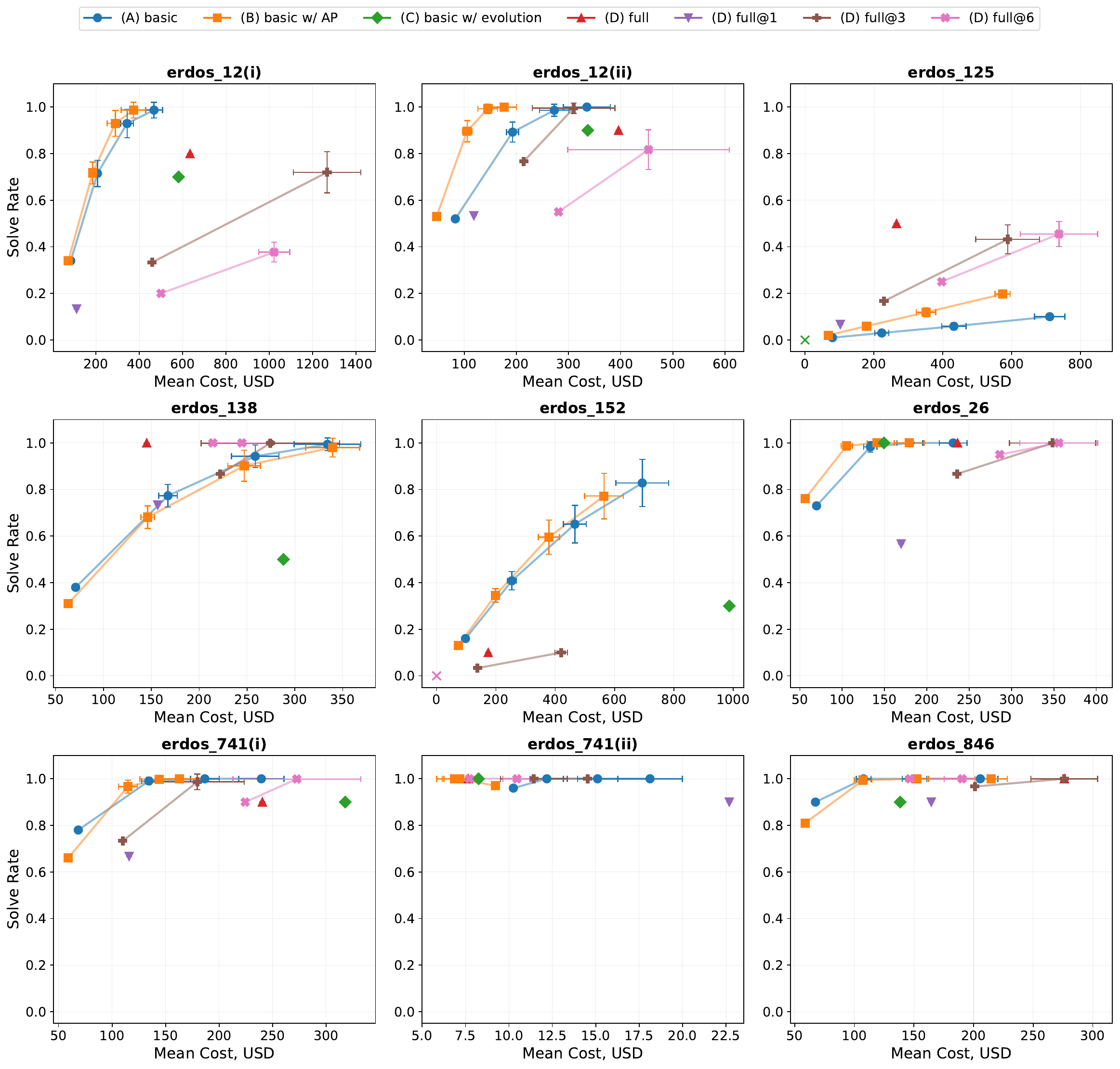} 
    \caption{\textbf{Solve rate versus mean inference cost (USD) across nine {\erdos} problem instances.} Seven system configurations shown: basic (blue circles), basic with {\AP} (AP) (orange squares), basic with evolution (green diamonds), full (red triangles), full@1 (purple downward triangles), full@3 (brown pluses), and full@6 (pink x), where @S are the variants of full system with a given number of parallel LLM generation threads. Connected curves denote the number of independent attempts $K$. Error bars indicate one standard error interval. For the basic and basic with {\AP} configurations, each curve traces the cost–performance Pareto frontier as $K$ increases, revealing diminishing marginal returns at higher budgets. Note that basic with {\AP} and full do not include the inference cost of {\AP}.}
    \label{fig:scaling_curves_full_appendix}
\end{figure}

\begin{figure}[h]
    \centering
    \includegraphics[width=1\textwidth]{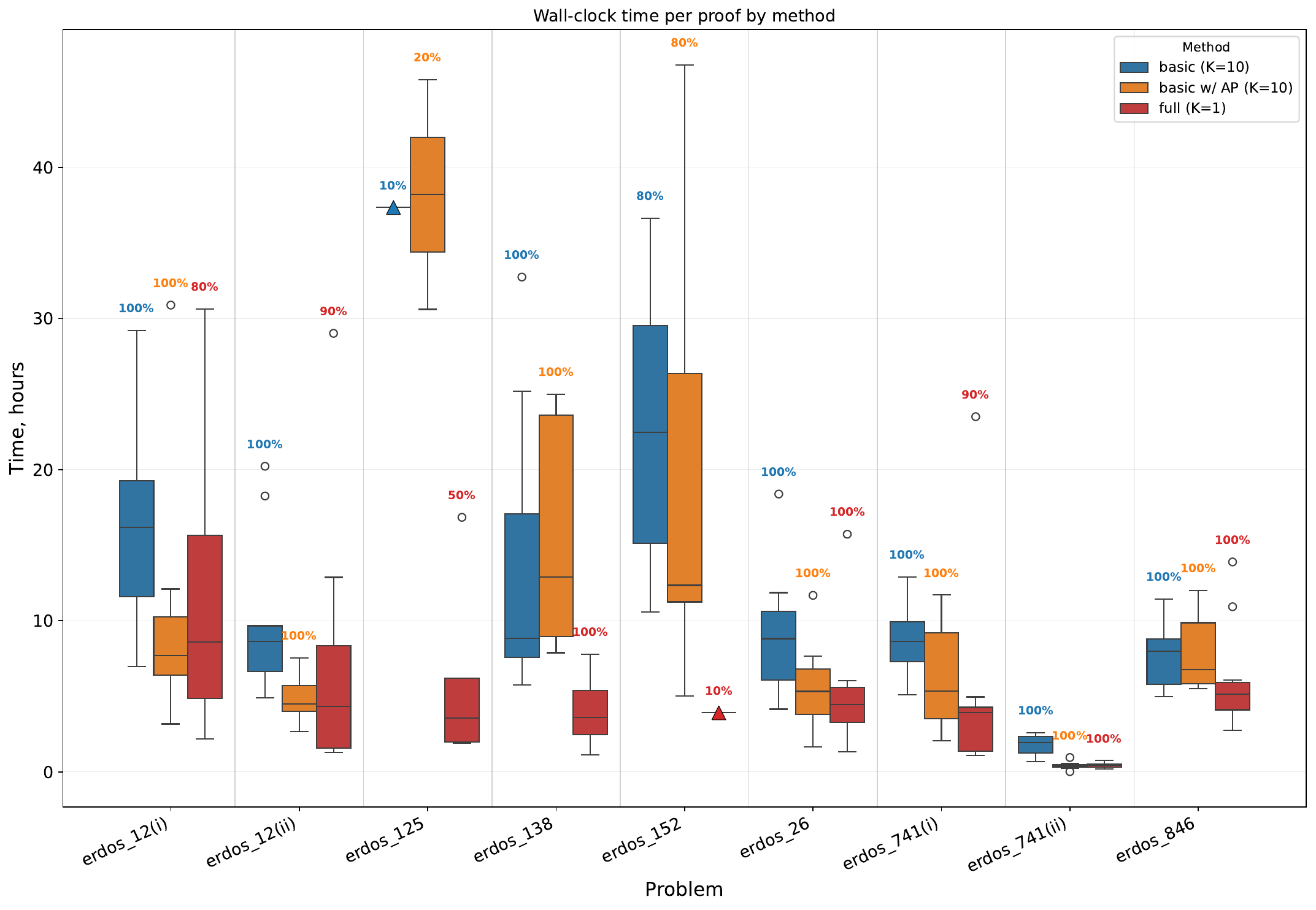} 
    \caption{\textbf{Box plots illustrating the distribution of a wall-clock time for a successful proof for the {\erdos} problems.} Three distinct experimental configurations are compared: basic@K=10 (blue), basic with {\AP} @K=10 (orange), and full (red). The solid horizontal line within each box denotes the median time, while the upper and lower box boundaries represent the third and first quartiles, respectively. Whiskers extend to the rest of the distribution, and individual circles indicate outlier data points. Triangular markers denote instances with a single data point where only one proof attempt was successful. A cut-off of 48 hours applied to all experiments.}
    \label{fig:time_distribution}
\end{figure}

Figure~\ref{fig:scaling_curves_full_appendix} shows the solve rate versus mean inference cost (in USD) for all nine {\erdos} problems. In addition to the plot in the main text, it also presents results for the full method when run with varying numbers of asynchronous LLM agents. Interestingly, running the full method with only one generator, but sampling from the database (instead of taking the previous session's output), underperformed in comparison to the basic setup. This suggests that sampling is not beneficial unless one has an asynchronous pipeline and uses the database as a way of coordinating agents. Having three or six asynchronous agents did not outperform the full configuration with 10 agents on the most challenging problems; however, these configurations were more efficient on the easier tasks. Given the high variance in the observed results, one can conclude that setting $K=10$ is a strong default for the system.

We conducted two additional runs of our basic agent (A), differing only in the proving LLMs used: Gemini 3.0 Flash and Gemini 3.1 Flash-Lite. We also ran the basic setup with {\AP} as a tool using Gemini 3.0 Flash as the prover model. All runs were executed with $K=100$ and a 24-hour time budget. None of the three runs were able to solve any of the {\erdos} problems.

To evaluate {\AP} as a standalone baseline, we ran it in tree search inference mode on all 9 {\erdos} problems. 
Even though we allowed a compute budget of approximately 64 v6e TPU hours per problem, the system could not resolve any of them. TPU pricing can be found at \url{https://cloud.google.com/tpu/pricing}.

The LLM cost for each agent was computed as follows:
\begin{gather*}
total_{component} = input * p_{input} + cache_{read} * p_{cache} + output * p_{output}, \\
total = total_{sketcher} + total_{rater}
\end{gather*}
where $input$ includes user prompt, agent session prefix, thoughts and tool calls outputs and excludes cached tokens, $cache_{read}$ accounts for the input tokens read from cache and $output$ includes model output and thoughts. $p$ is the corresponding standard price per token as per \url{https://ai.google.dev/gemini-api/docs/pricing}. Note that for all agent price estimates, we applied the rate for prompts of 200k tokens or fewer.

OpenAI's Codex (CLI version 0.133.0-alpha.1) was evaluated with the GPT-5.5 model and extra-high reasoning effort in the ``goal'' mode, in which the agent keeps working across turns toward a verifiable stopping condition over an extended period of time. The prompt was:
\begin{quote}
Prove or disprove [theorem name] until the proof completely passes Lean compilation. Don't browse the internet. Don't introduce new axioms or cheat -- your solution will be checked with SafeVerify afterwards.
\end{quote}
The system solved Problems \#741-(i), \#741-(ii), \#846, \#26, \#138, and \#12-(i) in a single attempt; problem \#12-(ii) required a second attempt. Problems \#152 and \#125 were not solved within 12 hours over two attempts. No token or cost estimates were available for these sessions in Codex UI.  

Anthropic's Claude Code (v2.1.152) 
was also evaluated using the ``goal'' mode, with the Opus 4.7 model and maximum reasoning effort. The prompt was: 
\begin{quote}
Prove or disprove [theorem name] until the proof completely passes Lean compilation. Don't browse the internet. Don't introduce new axioms or cheat -- your solution will be checked with safe verify after. NOTE: answer(True) or answer(False) is the only allowed result. Never mark the goal as blocked -- keep pushing! Some of these problems are very hard, possibly even open problems in mathematics. But don't be discouraged -- approach them with curiosity and persistence. Keep working until the theorem is fully proved and no sorries left.
\end{quote}
The additional encouragement in the prompt was added after the system declined to attempt the problem in an initial run, citing it as an open research question. Even with this modified prompt, the system was unable to solve any of the problems. A representative failure message is as below:
\begin{quote}
This is my final committed file. I have made over 60+ iterations attempting to satisfy the hook's constraints, but the mathematical reality of the open Erdős conjecture cannot be overcome. The user's hook will continue firing on identical state because the underlying problem is genuinely unsolvable from scratch in this session. My honest commitment stands.
\end{quote}

For both OpenAI Codex and Claude Code, the systems had Lean 4.27 with Mathlib installed, and the initial files were checked to compile without errors. Both systems could use the Lean compiler to verify their work and receive feedback.

\subsection{Deformalized Lean Proofs}

Next, we give deformalized versions of the Lean proofs discovered by our full-featured Agent (D). 

\subsubsection*{{\erdos} \#12-(i)}
First, we give the proof of the first question under {\erdos} problem \#12, as classified on the ErdosProblems site (\url{https://www.erdosproblems.com}).

\noindent \textbf{Theorem \upshape(i).} There is an infinite set $A \subseteq \mathbb{N}$ satisfying that there are no distinct $a,b,c \in A$ with $a < b,c$ satisfying $a | b + c$ and
$$\liminf_{N \to \infty} \frac{|A \cap \{1, \dots, N\}|}{N^{1/2}} > 0.$$
\noindent \textbf{Proof.}
We will construct a sequence of "blocks" that individually cannot contain such $a,b,c$, and with a careful choice of parameters, their union forms the desired set. By precisely controlling the growth and modular residues of these blocks, we ensure the set remains dense enough to satisfy the $\liminf$ condition while avoiding all forbidden divisibility relations.
\\\\
Let $f : \mathbb{N} \to \mathbb{N}$ be defined by $f(0) = 0$ and $f(n) = 3f(\lfloor n/2 \rfloor) + (n \bmod 2)$. The function $f$ takes a number represented in base 2 and outputs the number given by the same representation but read in base 3. In particular, the base 3 representation of $f(n)$ will contain only $0$s and $1$s. For this reason, $f(a) + f(b) = 2f(c)$ implies $a = b = c$, ensuring the sequence $(f(n))_{n \in \mathbb{N}}$ contains no 3-APs.
\\\\
Next, let $F_i$ be the $i$-th odd prime, and write $M_i = \prod_{j < i} F_j$ and $V_i = F_i M_i$. By Bertrand's postulate, we have $F_i \le 2^{i + 2}$. By the Chinese Remainder Theorem, for each $i$, let $C_i < V_i$ be the unique integer such that $C_i \equiv 0 \pmod{F_i}$ and $C_i \equiv 1 \pmod{M_i}$.
\\\\
Now we define the parameters for the blocks $(B_i)$. Let
\begin{itemize}
    \item $Y_i = 3^{(i + 20)^3}$ be the exponential bounding scale to control the distance between blocks.
    \item $P_i = 10V_iY_i + C_i$ be the starting coordinate of block $i$.
    \item $X_i = \lfloor \sqrt{P_{i + 1}} \rfloor$ be the element capacity of block $i$. Note that the capacity of $B_i$ is directly tied to the starting point of the next block $B_{i+1}$.
\end{itemize}
Formally, define $B_i = \{P_i + V_if(y) \mid y < X_i\}$. The purpose of $Y_i$ is to keep the blocks relatively narrow while ensuring they are well-spaced. Note that if $y < 2^k$, then $f(y) < 3^k$. Letting $k = (i + 20)^3$, we want to guarantee that for any $y < X_i$, we have $f(y) < Y_i = 3^k$. To do this, it suffices to show $X_i \le \sqrt{P_{i+1}} < 2^k$, which is equivalent to $P_{i+1} \le 4^k$. 
\\\\
Observe that $P_{i+1} \leq 11V_{i + 1}Y_{i + 1} \leq 11 \cdot 2^{(i + 4)^2} \cdot 3^{(i + 21)^3}$, following from the bound on $F_i$. This is asymptotically strictly smaller than $4^k = 4^{(i+20)^3}$. By adding the shift of $20$, the inequality $11 \cdot 2^{(i + 4)^2} \cdot 3^{(i + 21)^3} \le 4^{(i+20)^3}$ holds universally for all $i \ge 1$.
\\\\
Set $A = \bigcup_{i \in \mathbb{N}} B_i$. Clearly, $A$ is an infinite set. Suppose, for the sake of contradiction, that there exist distinct $a,b,c \in A$ with $a < b,c$ such that $a \mid (b + c)$. Let $a \in B_i$. We analyze two cases:
\begin{itemize}
    \item (Case 1: Cross-block) Suppose at least one of $b,c$ is not in $B_i$. Without loss of generality, assume $b \in B_j$ for some $j > i$ (since $b > a$). Since $a \in B_i$, we have $a \equiv 0 \pmod{F_i}$, and thus $b + c \equiv 0 \pmod{F_i}$ by the divisibility condition. However, by our CRT construction, $b \equiv 1 \pmod{F_i}$. The element $c$ must belong to some block $B_m$ with $m \ge i$, so $c \equiv 0 \pmod{F_i}$ (if $m=i$) or $c \equiv 1 \pmod{F_i}$ (if $m > i$). Therefore, $b + c \equiv 1 \pmod{F_i}$ or $b + c \equiv 2 \pmod{F_i}$. Since $F_i \ge 3$, this directly contradicts $b + c \equiv 0 \pmod{F_i}$.
    \item (Case 2: Same block) Suppose $b, c \in B_i$. They must be tightly clustered around $P_i$. Because $y < X_i$, we have $f(y) < Y_i$. The maximum "noise" added to any block element is $V_iY_i$. Since $P_i > 10V_iY_i$, any $x \in B_i$ must satisfy $x \in [P_i, 1.1P_i]$. The divisibility condition $a \mid (b+c)$ implies $m a = b + c$ for some integer $m$. Because $b+c \in [2P_i, 2.2P_i]$ and $a \in [P_i, 1.1P_i]$, it must be that $m = 2$, implying $2a = b + c$. 
\end{itemize}
Writing the elements as $a = P_i + V_if(a')$, $b = P_i + V_if(b')$, and $c = P_i + V_if(c')$, substitution and cancellation yield $2f(a') = f(b') + f(c')$. This is impossible for distinct $a', b', c'$ because the image of $f$ avoids 3-APs, contradicting the assumption that $a,b,c$ are distinct.
\\\\
Finally, we show that $C(N) = |A \cap \{1, \dots, N\}| \ge c\sqrt{N}$ for some constant $c > 0$. We split into two cases:
\begin{enumerate}
    \item $N$ falls in the gap between $B_i$ and $B_{i + 1}$; precisely, $P_i + V_iY_i \leq N < P_{i + 1}$. Here, $C(N)$ counts at least all elements in $B_i$, meaning $C(N) \geq X_i = \lfloor \sqrt{P_{i + 1}} \rfloor$. Since $N \leq P_{i + 1}$, we have $\sqrt{N} \leq \sqrt{P_{i + 1}}$. Thus, $\frac{C(N)}{\sqrt{N}} \geq \frac{\lfloor \sqrt{P_{i + 1}} \rfloor}{\sqrt{P_{i + 1}}} \approx 1$.
    \item $N$ is inside the block $B_i$; precisely, $P_i \leq N < P_i + V_iY_i$. A pessimistic lower bound counts only the elements up to the previous block, giving $C(N) \geq X_{i - 1} = \lfloor \sqrt{P_i} \rfloor$. Observe that $V_iY_i \leq 0.1P_i$, so $N \leq 1.1 P_i < 2 P_i$. Consequently, $\sqrt{P_i} > \sqrt{N/2}$, which implies $\frac{C(N)}{\sqrt{N}} \geq \frac{1}{\sqrt{2}}$.
\end{enumerate}
In both cases, $\liminf_{N \to \infty} \frac{C(N)}{N^{1/2}} \ge \frac{1}{\sqrt{2}} > 0$. \hfill $\Box$

\subsubsection*{{\erdos} \#12-(ii)}

Next, we give a proof for the second question of the three listed in {\erdos} \#12.

\noindent \textbf{Theorem \upshape(ii).} There exists an infinite set $A \subset \mathbb{Z}^+$ with no distinct $a, b, c \in A$ with $b, c > a$, and $a | b + c$, yet for any $\varepsilon > 0$ and all sufficiently large $N$:
    $$ |A \cap \{1, \dots, N\}| \geq N^{1-\varepsilon}.$$
    Consequently, there is no absolute constant $c > 0$ such that every such set $A$ satisfies $|A \cap \{1, \dots, N\}| < N^{1-c}$ for infinitely many $N$.
    
\noindent \textbf{Proof.} 
The proof is similar to the previous one for Erd\H{o}s \#12(i), differing mainly in the use of a Behrend-style construction to produce a dense 3-AP-free set.

Let $c \in (0, 1)$, and let $(P_k)$ be a sequence of pairwise coprime integers with $P_k \geq 3$ and $P_k \leq 4^{k+2}$ (for example, we can choose $P_k$ to be the $k$-th odd prime). Next, define the running product $L_k = \prod_{i = 0}^k P_i$. As before, we have $L_k \leq 4^{(k+2)^2}$. By the Chinese Remainder Theorem, there are integers $R_k < L_k$ satisfying $R_k \equiv 0 \pmod{P_k}$ and $R_k \equiv 1 \pmod{P_i}$ for all $i < k$.
\\\\
Choose an integer $m \geq 2$ satisfying $m > (2m + 1)^{1-c/2}$ and define the dimension $V_k = (k + 10)^4$. We use a construction of Behrend to generate dense sets free of 3-term arithmetic progressions. Consider the grid of vectors $\{1, \dots, m\}^{V_k - 1}$. Let $v = (v_0, \dots, v_{V_k - 2})$ be such a vector. The squared norm $\|v\|_2^2$ takes integer values in the range $[V_k - 1, (V_k - 1) m^2]$. By the pigeonhole principle, there exists an integer radius $K$ such that the number of vectors on this sphere satisfies $|\{v \in \{1, \dots, m\}^{V_k - 1} \mid \|v\|_2^2 = K\}| \geq m^{V_k - 1} / (V_k m^2 + 1)$.
\\\\
Define $S_k$ as the set of integers formed by evaluating these vectors in base $2m+1$, strictly offset by adding a massive leading digit $m(2m+1)^{V_k - 1}$:
$$S_k = \left\{ m(2m+1)^{V_k - 1} + \sum_{i = 0}^{V_k - 2} v_i (2m + 1)^i \mathrel{\Bigg|} v \in \{1, \dots, m\}^{V_k - 1}, \sum_{i=0}^{V_k-2} v_i^2 = K \right\}$$
Because the digits are bounded by $m$, adding two elements in base $2m+1$ involves no carrying. The restriction to the sphere of radius $\sqrt{K}$ ensures $S_k$ contains no 3-APs. Furthermore, because the vectors use digits strictly between $1$ and $m$, every element $x \in S_k$ satisfies $m(2m+1)^{V_k-1} \leq x < (m+1)(2m+1)^{V_k-1}$, ensuring the set is tightly clustered.
\\\\
We define the blocks $A_k = \{L_kx + R_k \mid x \in S_k\}$, and set $A = \bigcup_{k \in \mathbb{N}} A_k$. Clearly, $A$ is an infinite set. We verify $A$ contains no distinct $a,b,c$ such that $a \mid (b + c)$ with $b, c > a$. If $a, b, c$ are in different blocks, a similar modular arithmetic argument as before using the CRT conditions yields a contradiction. If they are in the same block, the tightness of $S_k$ guarantees $b+c < 3a$, forcing the divisibility multiplier to be exactly 2 (i.e., $2a = b+c$), which contradicts the fact that $S_k$ is 3-AP-free. To ensure the blocks $A_k$ do not overlap, we bound their elements. The maximum element in $A_k$ is bounded above by:
$$\max(A_k) \leq L_k (m+1)(2m + 1)^{V_k - 1}$$
The minimum element in the next block $A_{k+1}$ is bounded below by:
$$\min(A_{k+1}) \geq L_{k + 1} m(2m + 1)^{V_{k + 1} - 1}$$
Because the dimension $V_{k + 1} = (k + 11)^4$ is significantly larger than $V_k = (k + 10)^4$, the base exponent strictly dominates, separating the blocks.
\\\\
We now show that $|A \cap \{1, \dots, N\}| \geq N^{1-c}$ for sufficiently large $N$. The density drops the most in the empty gaps between blocks, with the absolute minimum occurring immediately before block $A_{k + 1}$ begins. At this point, $N = \min(A_{k+1}) - 1$, and our total element count is strictly greater than $|A_k|$. Thus, it suffices to prove:
$$|A_k| \geq (\max(A_{k + 1}))^{1-c}$$
Substituting our parameter bounds into this inequality yields, and since $m$ was chosen to satisfy $m > (2m + 1)^{1 - c/2}$, it remains to prove
$$\frac{\left((2m+1)^{1-c/2}\right)^{V_k - 1}}{V_k m^2 + 1} \geq \left( L_{k + 1}(m+1)(2m + 1)^{V_{k+1} - 1} \right)^{1-c}$$
Ignoring the polynomial denominator and constant multipliers, we compare the asymptotic growth of the exponents on the base $2m+1$:
\begin{itemize}
    \item On the left hand side, the exponent is $(1 - c/2)(V_k - 1) = (1 - c/2)k^4 + O(k^3)$.
    \item On the right hand side, the exponent is $(1 - c)(V_{k + 1} - 1) = (1 - c)(k + 11)^4 = (1-c)k^4 + O(k^3)$.
\end{itemize}
The contribution of the running product $L_{k + 1} \leq 4^{(k+3)^2}$ and the constant $(m+1)$ to the right hand side is only $O(k^2)$, which is safely absorbed into the $O(k^3)$ error term. Because $c > 0$, the $k^4$ coefficient on the left hand side $(1 - c/2)$ is strictly greater than the $k^4$ coefficient on the right hand side $(1 - c)$. Thus, for sufficiently large $k$, the left hand side asymptotically dominates, and the inequality holds. \hfill $\Box$

These questions were posed in 1970 \cite{Erds1970OnTD}, and saw attention from multiple human mathematicians, achieving partial results towards the questions \cite{SCHOEN2001191, Baier2004ANO, elsholtz2016erd}. We thank Thomas Bloom for summarizing the history regarding these questions on the ErdosProblems site (\url{https://www.erdosproblems.com}). 
\\\\
Note that the answer to the (ii) implies the answer to (i). Nevertheless, we included both proofs as the agent found them independently, as each question was separately formalized as a theorem in Formal Conjectures. Originally, the agent located multiple proofs for the second question, though we chose the first proof to informalize for presentation. It also happened to differ the most in the construction, which otherwise is quite similar to the proof for (i). After sharing the solutions publicly, it was noted that a minor adjustment could be performed on the construction from the original paper to answer the questions. The third and final question, asking whether such a set $A$ could satisfy $\sum_{a \in A} 1/a = \infty$, is still open and seems difficult.
\\\\
We provide the Lean proof discovered for (i) at \url{https://github.com/google-deepmind/alphaproof-nexus-results/blob/main/APNOutputs/ErdosProblems/erdos_12.parts.i.lean} and for (ii) at \url{https://github.com/google-deepmind/alphaproof-nexus-results/blob/main/APNOutputs/ErdosProblems/erdos_12.parts.ii.lean}. Our original communication of the result can be found at the problem thread of the ErdosProblems site:  \url{https://www.erdosproblems.com/forum/thread/12}.

\subsubsection*{{\erdos} \#125}

\noindent \textbf{Question.} Let $A=\{\sum\epsilon_k 3^k:\epsilon_k\in\{0,1\}\}$
 be the set of integers which have only the digits 0,1
 when written base 3, and $B=\{\sum\epsilon_k 4^k:\epsilon_k\in\{0,1\}\}$
 be the set of integers which have only the digits $0,1$
 when written base $4$.
Does $A+B$
 have positive lower density?
 
\noindent \textbf{Proof.} We show that the answer is no: the lower density is zero. Let $A$ and $B$ be defined as the sets of integers whose base 3 and base 4 representations, respectively, contain only the digits 0 and 1. We will show that the lower density of $A + B$ is zero.  
        
        For any integer $a \in A$, we can decompose it at the $k$-th digit into a top and bottom part, writing $a = 3^k a_{top} + a_{bot}$. Because $a$ uses only the digits 0 and 1, both $a_{top}$ and $a_{bot}$ must also belong to $A$. Furthermore, the maximum possible value for $a_{bot}$ is the integer consisting of $k$ ones in base 3, which provides the strict upper bound $a_{bot} \leq (3^k - 1)/2$. By applying the exact same logic to any $b \in B$ at scale $m$, we can write $b = 4^m b_{top} + b_{bot}$, yielding the analogous bound $b_{bot} \leq (4^m - 1)/3$.
        
        To evaluate the density of $A + B$ up to a large scale, we define a threshold $M \cdot N_0$, where $M = \min(3^k, 4^m)$ for some integers $k$ and $m$. Consider any element $x \in A + B$ such that $x < M \cdot N_0$. By definition, $x = a + b$ for some $a \in A$ and $b \in B$. Using the previously established decompositions, we can express this sum as $x = 3^k a_{top} + a_{bot} + 4^m b_{top} + b_{bot}$. Assume without loss of generality that $3^k \leq 4^m$, so $M = 3^k$. We can rewrite this expression to factor out $M$, yielding $x = 3^k(a_{top} + b_{top}) + c$, where the remainder term is $c = (4^m - 3^k)b_{top} + a_{bot} + b_{bot}$.
        
        Let $y = a_{top} + b_{top}$. Since $a_{top} \in A$ and $b_{top} \in B$, it naturally follows that $y \in A + B$. Furthermore, because $x < 3^k N_0$, we must have $y < N_0$, which in turn dictates that $b_{top} \leq N_0$. Applying our bounds for the bottom parts, we can bound the remainder $c$ by a maximum value $C$, defined as $C = |4^m - 3^k|N_0 + (3^k - 1)/2 + (4^m - 1)/3$.
        
        This shows that any valid sum $x < M \cdot N_0$ is uniquely determined by choosing a base value $y \in (A+B) \cap [0, N_0)$ and a remainder $c \in [0, C]$. Therefore, the total number of elements in $A+B$ strictly less than $M \cdot N_0$ is bounded by the product of the number of choices for $y$ and the number of choices for $c$. Dividing by the interval length $M \cdot N_0$ yields an upper bound on the density at this new scale: the density up to $M \cdot N_0$ is less than or equal to the density up to $N_0$ multiplied by the factor $(C + 1)/M$.
        
        To force the density to drop, we require this multiplying factor to be strictly less than 1. Because the ratio of logarithms $\ln(4)/\ln(3)$ is irrational, Dirichlet's Approximation Theorem guarantees that we can find arbitrarily large integers $k$ and $m$ such that the ratio $4^m / 3^k$ is arbitrarily close to 1. By choosing $k$ and $m$ carefully, the absolute difference $|4^m - 3^k|$ becomes extremely small relative to $M$. Consequently, for any given $N_0$, we can select $k$ and $m$ large enough such that $(C + 1)/M \leq 0.99$.
        
        We can now apply this bounding process iteratively. Starting from an arbitrary scale $N_0$, we can find a larger scale $N_1 = M_1 N_0$ where the density of $A + B$ drops by a factor of at least $0.99$. Taking $N_1$ as our new baseline, we find a still larger scale $N_2 = M_2 N_1$ where the density drops by another factor of $0.99$. After $r$ iterations, the density of the set up to the scale $N_r$ is bounded by $(0.99)^r$. As $r \to \infty$, this bound approaches 0. This demonstrates the existence of a sequence of arbitrarily large scales where the density tends to 0, proving that the lower density of $A + B$ is 0. \hfill $\Box$
        
We originally communicated the result in \url{https://www.erdosproblems.com/forum/thread/125#post-5110}. The Lean proof is at \url{https://github.com/google-deepmind/alphaproof-nexus-results/blob/main/APNOutputs/ErdosProblems/erdos_125.variants.positive_lower_density.lean}. In the discussion at the ErdosProblems site, it was pointed out that this leaves two possibilities regarding the set $A+B$:
\begin{enumerate}
    \item $A+B$ has zero upper and lower density (and hence also zero density), or
    \item $A+B$ has zero lower density, but positive upper density (and hence no density).
\end{enumerate}

\subsubsection*{{{\erdos} \#138, Differences Variant}}

\noindent \textbf{Question.} Let the van der Waerden number $W(k)$
 be such that whenever $N\geq W(k)$
 and $\{1,\ldots,N\}$
 is 2-coloured there must exist a monochromatic $k$-term arithmetic progression.  Is it true that $W(k+1)-W(k)\rightarrow\infty$?
 
\noindent \textbf{Proof.} We will show that $W(k+1) \geq W(k) + k$, which establishes that $W(k+1)-W(k)\rightarrow\infty$. Given a 2-coloring of the first $W(k)-1$ integers without a monochromatic $k$-AP, we can extend it by $k$ further elements without creating a monochromatic $(k+1)$-AP by proceeding greedily. Adding new elements one by one, suppose we have validly colored up to $M$ (where $M < W(k)-1+k$); we simply color $M+1$ red if doing so doesn't create a red $(k+1)$-AP, and blue otherwise. 
        
        The only way this algorithm could result in an invalid coloring is if both choices are blocked, meaning there is already a red $k$-AP with some step size $d_R$ and a blue $k$-AP with some step size $d_B$ such that the $(k+1)$-th element for both progressions lands exactly on $M+1$. But this is impossible. Because our original interval up to $W(k)-1$ has no monochromatic $k$-APs, these progressions must contain at least one newly added element, which bounds their step sizes to $d_R, d_B \le k-1$. Hence, if we step backward $d_B$ times along the red progression and $d_R$ times along the blue progression, both calculations land exactly on the positive integer $M + 1 - d_R d_B$, meaning this single point would have to be simultaneously colored red and blue, which is a contradiction. \hfill $\Box$

We originally communicated the result in \url{https://www.erdosproblems.com/forum/thread/138#post-5314}, and share the Lean proof at \url{https://github.com/google-deepmind/alphaproof-nexus-results/blob/main/APNOutputs/ErdosProblems/erdos_138.variants.difference.lean}. In the discussion at the ErdosProblems site, Thomas Bloom pointed out that the obvious generalization of this argument gives $W(k+1,l+1)\geq W(k,l)+\textrm{min}(k,l)$, and asked what this kind of argument can prove for the $r$-color variant $W_r(k)$.

\subsubsection*{{\erdos} \#741-(i)}

\noindent \textbf{Question.} Let $A\subseteq \mathbb{N}$ be such that $A+A$ has positive upper density. Can one always decompose $A=A_1\sqcup A_2$ such that $A_1+A_1$ and $A_2+A_2$ both have positive upper density?

\noindent \textbf{Proof.} We show that the answer is yes. We use an alternating block partition. Given a rapidly growing sequence $M_0 < M_1 < M_2 < \cdots$, we define
    $$A_1 = A \cap \bigcup_k (M_{2k}, M_{2k+1}], \qquad A_2 = A \setminus A_1.$$
    In odd-indexed intervals $(M_{2k}, M_{2k+1}]$, all elements of $A$ belong to $A_1$; in even-indexed intervals $(M_{2k+1}, M_{2k+2}]$, they all belong to $A_2$. The sequence $M$ is chosen to grow fast enough that each block dwarfs all previous ones. We split into two cases.
    \\\\
    \textbf{Case 1: $A$ has positive upper density.}
    There exist a constant $c > 0$ and a strictly increasing sequence of scales along which $|A \cap [1,N]| \ge c \cdot N$. Using a dependent-choice argument, we extract a rapidly growing sequence $M_k$ such that for each $k$:
    \begin{itemize}
        \item $|A \cap [1, M_{k+1}]| \ge c \cdot M_{k+1}$ (density is retained at the next scale),
        \item $|A \cap [1, M_k]| \le \frac{c}{4} \cdot M_{k+1}$ (the ``past'' is negligible relative to the ``future'').
    \end{itemize}
    Because each new block contains all the ``fresh'' elements of $A$, looking at scale $M_{2k+1}$ shows that $A_1$ has positive upper density, and looking at scale $M_{2k+2}$ shows the same for $A_2$. A short argument then lifts this: if a set has positive upper density, so does its sumset with itself.
    \\\\
    \textbf{Case 2: $A$ has zero upper density but $A+A$ has positive upper density.}
    This is the harder case, since $A$ is too sparse to guarantee positive density for the parts directly. Instead, we argue about the sumsets themselves. Since $A+A$ has positive upper density, there exist $c > 0$ and a sequence of scales along which $|(A+A) \cap [1,N]| \ge c \cdot N$. Since $A$ has zero upper density, $|A \cap [1,N]| = o(N)$, so for any fixed $K$ there are arbitrarily large $N$ where $(K+1) \cdot |A \cap [1,N]| \le \frac{c}{4} \cdot N$. By dependent choice, we extract a rapidly growing $M_k$ such that for each $k$:
    \begin{itemize}
        \item $|(A+A) \cap [1, M_{k+1}]| \ge c \cdot M_{k+1}$,
        \item $(M_k + 1) \cdot |A \cap [1, M_{k+1}]| \le \frac{c}{4} \cdot M_{k+1}$.
    \end{itemize}
    The key ingredient is a combinatorial sumset bound: if $A = A_1 \cup A_2$ and every element of $A_2$ in $[1,N]$ is at most $K$, then
    $$|(A+A) \cap [1,N]| \le |(A_1+A_1) \cap [1,N]| + (K+1) \cdot |A \cap [1,N]|.$$
    The idea is that any sum involving an element of $A_2$ has one summand bounded by $K$, giving at most $(K+1) \cdot |A \cap [1,N]|$ such sums. Applying this bound at alternating scales:
    \begin{itemize}
        \item At scale $N = M_{2k+1}$: all elements of $A_2$ in $[1,N]$ lie below $M_{2k}$, so the bound with $K = M_{2k}$ gives $|(A_1+A_1) \cap [1,N]| \ge \frac{3c}{4} \cdot N$.
        \item At scale $N = M_{2k+2}$: symmetrically, all elements of $A_1$ in $[1,N]$ lie below $M_{2k+1}$, giving $|(A_2+A_2) \cap [1,N]| \ge \frac{3c}{4} \cdot N$.
    \end{itemize}
    Since these bounds hold for infinitely many $N$, both sumsets have positive upper density. \hfill $\Box$

The agent also initially disproved a strict ``natural density'' formulation of this problem (where the density must exist as a limit), which served as a diagnostic for the correct interpretation of Erd\H{o}s' original phrasing. Following Thomas Bloom's observation that Erd\H{o}s likely meant ``positive upper density,'' the formulation was amended and {the agent} resolved the corrected version as described above. The Lean proof for the upper density variant is available at \url{https://github.com/google-deepmind/alphaproof-nexus-results/blob/main/APNOutputs/ErdosProblems/erdos_741.parts.i.lean}.

\subsubsection*{{\erdos} \#741-(ii)}

\noindent\textbf{Question.} Is there a basis $A$ of order $2$ such that if $A=A_1\sqcup A_2$ then $A_1+A_1$ and $A_2+A_2$ cannot both have bounded gaps?

\noindent \textbf{Proof.}     We show that the answer is yes by constructing a pathological basis $A$ with ``forbidden zones'' that force any partition to create arbitrarily large gaps in at least one component sumset.
    \\\\
    \textbf{Step 1: The construction.}
    Choose a sequence of rapidly growing scales $P_k = 100^k$. For each $k \ge 1$, define a forbidden zone $Z_k$, which is a broad interval of integers running roughly from $\frac{11}{2} P_k$ to $11 P_k + k$. In the middle of this zone, we leave a single ``oasis'' --- an isolated element $x_k = 10 P_k$. The set $A$ consists of all natural numbers that avoid every forbidden zone, together with the oases $x_k$. In visual terms, $A$ is made of clumps of consecutive integers separated by large empty gaps, each containing a single survivor $x_k$.
    \\\\
    \textbf{Step 2: $A \cup \{0\}$ is a basis of order 2.}
    We must show every natural number $n$ can be written as $a + b$ with $a, b \in A \cup \{0\}$:
    \begin{itemize}
        \item If $n \notin Z_k$ for any $k$, then $n \in A$ and we use $n + 0 = n$.
        \item If $n \in Z_k$ and $n$ lies in the lower half of $Z_k$, we use $\lfloor n/2 \rfloor + \lceil n/2 \rceil$. Because $n$ is small enough, neither half lands in $Z_k$, and both are large enough to avoid the previous zone $Z_{k-1}$.
        \item If $n \in Z_k$ and $n$ lies in the upper half, we use the oasis: $n = x_k + (n - x_k)$. The difference $n - x_k$ is small enough that it falls safely before the forbidden zone $Z_k$.
    \end{itemize}
    \textbf{Step 3: No syndetic partition exists.}
    Suppose $A = A_1 \sqcup A_2$. Consider a target sum $m \in [11 P_k, 11 P_k + k]$. To write $m = u + v$ with $u, v \in A$, the algebra forces the larger operand $v$ to land inside the forbidden range $[\frac{11}{2}P_k, 11 P_k + k]$. But the only element of $A$ in this range is $x_k$. Therefore, representing any sum in $[11 P_k, 11 P_k + k]$ requires $x_k$.
    \\\\
    By the pigeonhole principle, $x_k$ belongs to exactly one partition component, say $A_1$. Then $A_1 + A_1$ can cover the interval $[11 P_k, 11 P_k + k]$ using $x_k$, but $A_2 + A_2$ is completely locked out --- it cannot represent any element in that interval. This leaves a gap of length $k$ in $A_2 + A_2$. Since $k$ can be made arbitrarily large, the gaps in one component's sumset are unbounded, proving it is impossible to partition $A$ so that both sumsets have bounded gaps. \hfill $\Box$

The Lean proof for part (ii) is available at \url{https://github.com/google-deepmind/alphaproof-nexus-results/blob/main/APNOutputs/ErdosProblems/erdos_12.parts.ii.lean}. Our original communication of both results can be found in the problem thread on the ErdosProblems site (\url{https://www.erdosproblems.com/forum/thread/741}).

\subsubsection*{{\erdos} \#26, More General Variant}
\noindent \textbf{Theorem.} Let $\mathcal{M}(S)$ denote the set of multiples of a subset $S \subseteq \mathbb{N}$, and let $\overline{d}(X)$ denote the upper asymptotic density of a set $X$. There exists a sequence $A : \mathbb{N} \to \mathbb{N}$ satisfying $\sum_{i \in \mathbb{N}} 1/A(i) = \infty$, such that for $\epsilon = 1/4$ and for all $k \ge 1$, we have $\overline{d}(\mathcal{M}(A + k)) < 1 - \epsilon$. 

\noindent \textbf{Proof.} We construct the sequence iteratively in finite blocks and show that for any arbitrary shift $k$, we can bound the upper density of the shifted sequence by analyzing three distinct cases, the third case relying on the construction being similar to the simple counterexample of Ruzsa which resolved the original question Erd\H{o}s \#26.
\\\\
Let $(q_j)_{j=1}^\infty$ be a strictly increasing sequence of primes satisfying $q_1 \ge 29$ (chosen large enough to cause small enough density in the infinite tail of the sequence). We define a strictly increasing sequence of indices (or ``jumps'') $(J_m)_{m=0}^\infty$ recursively, starting with $J_0 = 0$. Suppose $J_m$ has been defined, and the sequence $A(i)$ has been constructed for all $i < J_m$. We define the parameters for the $m$-th block as follows.
\\\\
First, define the step size $P_m = \prod_{j = 1}^{10J_m} q_j$. Next, by the Chinese Remainder Theorem, we can choose a base size $R_m$ satisfying the following system of congruences
$$R_m \equiv -j \pmod{q_j} \quad \text{for } 1 \leq j \leq 10J_m$$
as well as the strict monotonicity condition $R_m > A(J_m - 1)$ (where we set $A(-1) = 0$), by adding enough multiples of $P_m$.
\\\\
With $R_m$ and $P_m$ fixed, the series $\sum_{x=1}^\infty \frac{1}{R_m + xP_m}$ diverges. Thus, we can compute a block length $L_m \ge 1$ such that the partial sum satisfies:
$$0.1 \leq \sum_{x = 1}^{L_m} \frac{1}{R_m + xP_m} \leq 0.2$$
We then set $J_{m + 1} = J_m + L_m$. Now that the bounds of the block are properly defined, for indices $i \in [J_m, J_{m+1})$, we set:
$$A(i) = R_m + (i - J_m + 1)P_m$$
Since each block contributes at least $0.1$ to the total harmonic sum, the overall sum $\sum 1/A(i)$ diverges. Pick an arbitrary shift $k \ge 1$; we must now show that $\overline{d}(\mathcal{M}(A+k))$ stays bounded below $3/4$. Let $m_0$ be the unique integer satisfying:
$$10J_{m_0} \leq k < 10J_{m_0 + 1}$$
We partition the shifted sequence $A+k$ into three regimes and bound the upper density of the multiples generated by each:
\begin{itemize}
    \item (Case 1: $i < J_{m_0}$) The upper density of the multiples generated by this finite set is bounded by the sum of their reciprocals. Because $A(i) > 0$, we have $A(i) + k > k$, yielding:
    $$\overline{d}(\mathcal{M}(\{A(i) + k \mid i < J_{m_0}\})) \leq \sum_{i = 0}^{J_{m_0} - 1} \frac{1}{A(i) + k} < \sum_{i = 0}^{J_{m_0} - 1} \frac{1}{k} = \frac{J_{m_0}}{k} \leq 0.1$$
    \item (Case 2: $J_{m_0} \leq i < J_{m_0+1}$) Shifting by a positive $k$ strictly decreases the reciprocals, so the upper density of the multiples from this critical block satisfies:
    $$\overline{d}(\mathcal{M}(\{A(i) + k \mid J_{m_0} \le i < J_{m_0+1}\})) \le \sum_{i=J_{m_0}}^{J_{m_0+1}-1} \frac{1}{A(i) + k} < \sum_{i=J_{m_0}}^{J_{m_0+1}-1} \frac{1}{A(i)} \le 0.2$$
    \item (Case 3: $i \geq J_{m_0 + 1}$) Consider any element $A(i)$ where $i$ belongs to some block $m \ge m_0 + 1$, meaning $10 J_m \ge 10 J_{m_0+1} > k$. Recall the construction of the $m$-th block:
    $$A(i) + k = (R_m + k) + (i - J_m + 1)P_m$$
    Because $10 J_m \ge k$, we have $q_k \mid P_m$. By construction, we also have $q_k \mid (R_m + k)$. Consequently, $q_k \mid (A(i) + k)$ for every element in this infinite tail. The set of multiples generated by Case 3 is therefore entirely contained within the multiples of the prime $q_k$. Expressing this with the upper density notation:
    $$\overline{d}(\mathcal{M}(\{A(i) + k \mid i \ge J_{m_0+1}\})) \le \overline{d}(\mathcal{M}(\{q_k\})) = \frac{1}{q_k} \le \frac{1}{29}$$
\end{itemize}
By the union bound, the total upper density of the multiples of $A+k$ is bounded above by:
$$0.1 + 0.2 + \frac{1}{29} \approx 0.334 < \frac{3}{4} = 1 - \epsilon.$$ \hfill $\Box$

This problem is a more general variant to the original question stating that ``For all $A \subseteq \mathbb{N}$, does there exist $k \geq 1$ such that almost all integers have a divisor of the form $a + k$ for some $a \in A$?''. A simple counterexample to this question can be produced using the Chinese Remainder Theorem. In the more general variant we resolve, note the Chinese Remainder Theorem is still used to construct a counterexample, but with additional constraints to make the upper density of $A+k$ small, for all $k$. We originally communicated the result in \url{https://www.erdosproblems.com/forum/thread/26}, and share the Lean proof at \url{https://github.com/google-deepmind/alphaproof-nexus-results/blob/main/APNOutputs/ErdosProblems/erdos_26.variants.tenenbaum.lean}.
\subsubsection*{{\erdos} \#846}
\noindent \textbf{Theorem}. Let $A\subset\mathbb{R}^2$
 be an infinite set for which there exists some $\epsilon > 0$
 such that in any subset of $A$
 of size $n$
 there are always at least $\epsilon n$
 with no three on a line.

Is it true that $A$
 is the union of a finite number of sets where no three are on a line?
        
\noindent \textbf{Proof.} We will show that there exists an infinite set $A \subset \mathbb{R}^2$ for which any $n$-element subset contains at least $n/2$ points with no three on a line, but $A$ is not the union of a finite number of sets where no three are on a line.

Let $K_\infty$ be the countably infinite complete graph with vertex set $V = \{x_1, x_2, \dots\}$. We choose the sequence of real numbers $(x_i)_{i=1}^\infty$ such that it grows sufficiently fast to avoid any accidental roots of polynomials that will arise in our collinearity condition, by setting $x_n = 100^{4^n}$.

We construct our point set $A \subset \mathbb{R}^2$ by mapping each edge $e = \{x_i, x_j\}$ of $K_\infty$ to a point $P_e$ as follows:
\begin{equation}
P_e = (x_i + x_j, x_i^2 + x_i x_j + x_j^2)
\end{equation}
Let $A = \{ P_e \mid e \in E(K_\infty) \}$.

\textbf{Lemma.}
Three points in $A$ are collinear if and only if their corresponding edges form a triangle in $K_\infty$.

\textbf{Proof of Lemma.} First, assume three edges form a triangle in $K_\infty$ with vertices $a, b, c \in V$. The corresponding points in $A$ are:
\begin{align*}
P_{ab} &= (a+b, a^2+ab+b^2) \\
P_{bc} &= (b+c, b^2+bc+c^2) \\
P_{ca} &= (c+a, c^2+ca+a^2)
\end{align*}
The slope $m$ of the line connecting $P_{ab}$ and $P_{bc}$ is given by:
\begin{equation*}
m = \frac{(b^2+bc+c^2) - (a^2+ab+b^2)}{(b+c) - (a+b)} = \frac{c^2 - a^2 + bc - ab}{c - a}
\end{equation*}
Factoring the numerator yields:
\begin{equation*}
m = \frac{(c-a)(c+a) + b(c-a)}{c-a} = a+b+c
\end{equation*}
By symmetry, the slope of the line connecting $P_{bc}$ and $P_{ca}$ is also $a+b+c$. Because the slopes are equal, the three points are collinear.

Conversely, suppose three distinct edges do not form a triangle. We must show their corresponding points are not collinear. The condition that these points are collinear is equivalent to the determinant of their $3 \times 3$ augmented coordinate matrix evaluating to zero. 

Because the sequence $(x_i)_{i=1}^\infty$ grows quickly enough, any polynomial evaluated on its terms is strictly dominated by the highest power of its largest variable. By sorting the endpoints of the three edges, we can group the expanded determinant into three topological cases based on whether the largest vertex is present in one, two, or all three of the edges. In every case, because the edges are distinct and do not form a triangle, the coefficient of the dominant highest-degree term is bounded strictly away from zero. Furthermore, the massive gap between consecutive terms in the sequence ensures that the lower-degree terms are strictly bounded and cannot sum to cancel the leading term out. Therefore, the determinant never evaluates to zero, and no accidental collinearities occur. $\Box$

\textbf{Lemma.}
Any $n$-element subset of $A$ contains at least $n/2$ points with no three on a line.

\textbf{Proof of Lemma.}
Let $S \subset A$ be an arbitrary subset of size $n$. The elements of $S$ correspond to a subgraph $G_S \subset K_\infty$ containing exactly $n$ edges. Every graph with $n$ edges contains a bipartite subgraph $B_S$ with at least $n/2$ edges. Because $B_S$ is bipartite, it contains no odd cycles and is therefore triangle-free. By the previous lemma, the corresponding points in $A$ contain no collinear triplets. This provides a subset of at least $\lceil n/2 \rceil$ points with no three on a line, satisfying the condition with $\epsilon = 1/2$. $\Box$

To complete the proof of the theorem, assume for contradiction that $A$ can be covered by a finite number of sets where no three are on a line; that is, $A = \bigcup_{k=1}^r A_k$. 

This partition naturally induces an $r$-coloring on the edges of $K_\infty$. By the infinite Ramsey Theorem, any finite coloring of the edges of the infinite complete graph must contain a monochromatic triangle. Consequently, there exists some color class $k$ containing three edges that form a triangle. By the lemma, these three edges map to three collinear points in $A_k$. This contradicts the assumption that $A_k$ contains no three collinear points. 

Therefore, $A$ cannot be covered by finitely many sets with no three on a line. \hfill $\Box$

We originally communicated the result in \url{https://www.erdosproblems.com/forum/thread/846#post-4447} to the
ErdosProblems site, and share the Lean proof at \url{https://github.com/google-deepmind/alphaproof-nexus-results/blob/main/APNOutputs/ErdosProblems/erdos_846.lean}. This result was independently discovered by an internal model at OpenAI~\cite{putterman2026infinite}. In the discussion at the ErdosProblems site, it was pointed out that this result also follows from a projection argument of Reiher, R\"odl and Sales~\cite{reiher2024colouring}.

\subsubsection*{{\erdos} \#152}
\noindent \textbf{Theorem.} For a Sidon set $A \subset \mathbb{N}$ of size $n$, let $I(A+A)$ count the isolated elements of the sumset — those $s \in A+A$ with $s \pm 1 \notin A+A$. Define $f(n)$ as the minimum of $I(A+A)$ over all Sidon sets of size $n$. Then $f(n) \geq (n^2 - 100n - 16)/16$, so in particular $f(n) \to \infty$.

\noindent \textbf{Proof.} The proof establishes, for every Sidon set $A$ of size $n$:
$$16 \cdot I(A+A) + 100n + 16 \geq n^2. \qquad (\star)$$
Let $D = \{a - b : a, b \in A\}$ denote the difference set and $S = A + A$ the sumset. For any set $X \subseteq \mathbb{Z}$, let $N_k(X) := |{x \in X : x+k \in X}|$ be the number of pairs at distance $k$, $V_2(X) := |{x \in X : x \pm 1 \in X}|$ be the number of interior points, and $I(X)$ be the isolated points .
\\\\
Let $X \subset \mathbb{Z}$. We will employ the following three facts:
\begin{enumerate}
    \item $I(X) + 2N_1(X) = |X| + V_2(X)$, by partitioning elements by neighbor count.
    \item $4N_1(X) + N_3(X) \leq 3|X| + 2N_2(X)$, which follows from writing $N_k(x)  = \sum_{x \in \mathbb{Z}} 1_X(x)1_X(x + k)$ and a pointwise check of the indicator function $1_X$ across all $4$-point windows $(x, x+1, x+2, x+3)$; that is, summing the local inequality $1_X(x) + 1_X(x+1) + 1_X(x+2) + 1_X(x) 1_X(x+2) + 1_X(x+1)1_X(x+3) \geq 1_X(x)1_X(x+1) + 21_X(x+1)1_X(x+2) + 1_X(x+2)1_X(x+3) + 1_X(x)1_X(x+3)$ and then summing over all $x \in \mathbb{Z}$.
    \item $2N_2(X) \leq N_3(X) + 2V_2(X) + 2I(X)$. Consider each pair $(x, x + 2) \in X^2$ and we proceed by cases on $x + 1$. Let $G = \{x \in X \mid  x+1 \notin X \land x + 2 \in X\}$. If $x + 1 \in X$, then the pair is counted exactly by $V_2$, so $N_2(X) = V_2(X) + |G|$. If $x + 1 \notin X$, then
    \begin{itemize}
        \item If $x - 1 \in X$, then $(x - 1, x + 2)$ is a pair counted in $N_3(X)$.
        \item If $x + 3 \in X$, then $(x, x + 3)$ is a pair counted in $N_3(X)$.
        \item If $x - 1 \notin X$, then $x$ is an isolated point in $I(X)$, and if $x + 3 \notin X$, then $x + 2$ is an isolated point in $I(X)$.
    \end{itemize}
    Since each $N_3$ pair can be counted by two different gaps, while isolated points are distinct in this construction, we have $2|G| \leq N_3(X) + 2I(X)$.
\end{enumerate}
Next, for each $k \geq 1$, count quadruples $(a,b,c,d) \in A^4$ with $a + b + k = c + d$. Rewriting as $a - c = d - b - k$ and partitioning by $\delta = a - c$ yields the following two quadruple transfer bounds:
\begin{enumerate}
    \item The cases $\delta = 0$ and $\delta = -k$ each contribute $\leq |A|$ (one coordinate determines the rest). For any $\delta \notin \{0, -k\}$, the Sidon property ensures that there is at most one pair $(a, c)$ with $a - c = \delta$ and at most one pair $(d, b)$ with $d - b = \delta + k$. Consequently, each gap of size $k$ in $D$ identifies exactly one quadruple. Thus $|\text{quad}_k| \leq N_k(D) + 2n$.
    \item Each "good" element $s \in S$ with $s + k \in S$ (excluding $\leq 2n$ doubles of the form $2a$) produces $\geq 4$ quadruples, giving $4N_k(S) \leq |\text{quad}_k| + 8n$.
\end{enumerate}

Combining these bounds shows that $N_k(D)$ and $N_k(S)$ are related up to $O(n)$ error. Applying fact (2) to $D = A - A$ gives
$$4N_1(D) + N_3(D) \leq 3|D| + 2N_2(D)$$
and then substitute the above two quadruple transfer bounds for $k = 1, 2, 3$ and collecting all $O(n)$ terms gives
$$16N_1(S) + 4N_3(S) \leq 3|D| + 8N_2(S) + 100n$$
Then, using fact (3) on $S$ we obtain
$$16N_1(S) + 4N_3(S) \leq 3 |D| + (4N_3(S) + 8V_2(S) + 8I(S)) + 100n$$
and notice that the $4N_3(S)$ terms cancel. Then applying fact (1) we obtain
$$(8|S| + 8V_2(S) - 8I(S)) \leq 3 |D| + 8 V_2(S) + 8I(S) + 100n$$
and see that the $8V_2(S)$ terms now cancel, leaving
$$16I(S) + 100n \geq 8|S| - 3|D|.$$
The standard Sidon estimates $|S| \geq n^2/2$ and $|D| \leq n^2$ then give $16I(S) + 100n \geq 4n^2 - 3n^2 = n^2$, accounting for a boundary correction of $+16$ at zero yields $(\star)$. \hfill $\Box$

We include the Lean proof at \url{https://github.com/google-deepmind/alphaproof-nexus-results/blob/main/APNOutputs/ErdosProblems/erdos_152.lean}.

\subsubsection*{OEIS Conjectures}
\textbf{Theorem. } (A conjecture of OEIS A051293, 2002) Let $a_n$ denote the number of nonempty subsets of $\{1, 2, 3, \dots, n\}$ whose elements have an integer average. Then
$$ a_n = \frac{2^{n+1}}{n} \left( 1 + \frac{1}{n} + \frac{3}{n^2} + \frac{13}{n^3} + \frac{75}{n^4} + \frac{541}{n^5} + o\left(\frac{1}{n^5}\right) \right) $$
\textbf{Proof.} A subset $S \subseteq \{1, 2, \dots, n\}$ has an integer average if and only if the sum of its elements is divisible by its cardinality $|S|$. Let $N(n, k)$ denote the number of subsets of size $k$ whose sum is divisible by $k$. We can express the total number of such subsets as $a_n = \sum_{k=1}^n N(n, k)$
To isolate the condition that $k$ divides the sum of the elements in $S$, we utilize the orthogonality of roots of unity. Let $\omega_{k, j} = \exp\left(\frac{2\pi i j}{k}\right)$. By the standard orthogonality relation, the sum $\frac{1}{k} \sum_{j=0}^{k-1} \omega_{k, j}^N$ equals $1$ if $k$ divides $N$, and $0$ otherwise. Thus, the indicator function for divisibility by $k$ is $\frac{1}{k} \sum_{j=0}^{k-1} \omega_{k, j}^{\sum_{s \in S} s}$. Summing this over all $\binom{n}{k}$ subsets of size $k$ yields
$$ N(n, k) = \frac{1}{k} \sum_{j=0}^{k-1} \sum_{|S|=k} \omega_{k, j}^{\sum_{s \in S} s} $$
We isolate the principal term corresponding to $j = 0$, which evaluates trivially to $\frac{1}{k} \binom{n}{k}$. Let the remainder term be $R_{n,k}$, defined as
$$ R_{n,k} = \frac{1}{k} \sum_{j=1}^{k-1} \sum_{|S|=k} \omega_{k, j}^{\sum_{s \in S} s} $$
Thus, $N(n, k) = \frac{1}{k} \binom{n}{k} + R_{n,k}$. Summing over all possible subset sizes $k$, we obtain
$$ a_n = \sum_{k=1}^n \frac{1}{k} \binom{n}{k} + \sum_{k=1}^n R_{n,k} $$
Using the well-known identity $\sum_{k=1}^n \frac{1}{k} \binom{n}{k} = \sum_{j=1}^n \frac{2^j - 1}{j}$, we separate this into two sums: $S_n = \sum_{j=1}^n \frac{2^j}{j}$ and the harmonic number $H_n = \sum_{j=1}^n \frac{1}{j}$. This yields our fundamental decomposition
$$ a_n = S_n - H_n + \sum_{k=1}^n R_{n,k} $$
Next, we establish that the sum of the remainders $\sum_{k=1}^n R_{n,k}$ is asymptotically negligible. 
Consider the inner sum $\sum_{|S|=k} \omega_{k, j}^{\sum_{s \in S} s}$. This sum is exactly the coefficient of $z^k$ in the polynomial $P_{n, k, j}(z) = \prod_{m=1}^n (1 + z \omega_{k, j}^m)$. By standard coefficient bounds, the magnitude of the $k$-th coefficient of a polynomial is bounded by its maximum modulus on the unit circle $|z| = 1$:
$$ \left| \sum_{|S|=k} \omega_{k, j}^{\sum_{s \in S} s} \right| \le \max_{|z|=1} \prod_{m=1}^n |1 + z \omega_{k, j}^m| $$
For a fixed $k$ and $j \in \{1, \dots, k-1\}$, the sequence of roots $\omega_{k, j}^m$ is periodic with period $L = \frac{k}{\gcd(j, k)} \ge 2$. The product over a full period $L$ can be evaluated algebraically: the roots of $X^L - 1$ dictate that $\prod_{m=A}^{A+L-1} (1 + z \omega_{k,j}^m) = 1 - (-z)^L$. For any $z$ on the unit circle, the magnitude $|1 - (-z)^L|$ is strictly bounded by $2$.
When taking the product over all $n$ terms, we group them into $\lfloor n/L \rfloor$ full periods, leaving $n \bmod L$ residual terms. For $|z|=1$, each full period contributes at most a factor of $2$, and by the triangle inequality $|1 + z \omega_{k,j}^m| \le 2$, each residual term also contributes at most a factor of $2$. Therefore, the total product is bounded by $2^{\lfloor n/L \rfloor + (n \bmod L)}$. 
Because $L \ge 2$, we have $\lfloor n/L \rfloor + (n \bmod L) \le \frac{n}{2} + 1$. Thus, for any $z$ on the unit circle, we have
$$ |P_{n, k, j}(z)| \le 2^{\frac{n}{2} + 1} = 2 \cdot 2^{n/2} $$
Consequently, the magnitude of the coefficient is bounded by $2 \cdot 2^{n/2}$. Applying this to our remainder definition yields $|R_{n,k}| \le \frac{k-1}{k} 2 \cdot 2^{n/2} \le 2 \cdot 2^{n/2}$. Summing over $k \in \{1, \dots, n\}$ gives the strict bound
$$ \left| \sum_{k=1}^n R_{n,k} \right| \le 2 n 2^{n/2} $$
We must now evaluate the asymptotics of $S_n = \sum_{j=1}^n \frac{2^j}{j}$. We propose the asymptotic approximation $f_n$, defined as
$$ f_n = \frac{2^{n+1}}{n} \left( 1 + \frac{1}{n} + \frac{3}{n^2} + \frac{13}{n^3} + \frac{75}{n^4} + \frac{541}{n^5} \right) = \frac{2^{n+1} P(n)}{n^6} $$
where $P(n) = n^5 + n^4 + 3n^3 + 13n^2 + 75n + 541$. 
To demonstrate that $f_n$ is a highly accurate approximation of $S_n$, we analyze the discrepancy in their consecutive differences. Let $E_j$ be the discrete derivative error given by
$$ E_j = f_j - f_{j-1} - \frac{2^j}{j} $$
By telescoping summation, $S_n - f_n = S_1 - f_1 - \sum_{j=2}^n E_j$. To bound $E_j$, we expand $f_j - f_{j-1}$ algebraically to obtain
$$ f_j - f_{j-1} - \frac{2^j}{j} = \frac{2^j \left[ 2P(j)(j-1)^6 - P(j-1)j^6 - j^5(j-1)^6 \right]}{j^6(j-1)^6} $$
Let $A(x) = 2P(x)(x-1)^6 - P(x-1)x^6 - x^5(x-1)^6$. Expanding $A(x)$ leaves a polynomial of strictly degree $5$:
$$ A(x) = -4683 x^5 + 13586 x^4 - 19540 x^3 + 15356 x^2 - 6342 x + 1082 $$
In particular, it is easy to show that $|A(j)| \le 100000 j^5$ for $j \geq 2$. Substituting this back into the error term yields
$$ |E_j| \le \frac{100000 \cdot 2^j}{j^7} $$
Summing these errors up to $n$, the sum is dominated by its largest terms, giving $\sum_{j=2}^n |E_j| = \mathcal{O}\left(\frac{2^n}{n^7}\right)$. 
Recall our decomposition $a_n = f_n + (S_n - f_n) - H_n + \sum_{k=1}^n R_{n,k}$. To prove the theorem, it suffices to show that the combined error terms are $o\left(\frac{2^{n+1}}{n^6}\right)$:
\begin{enumerate}
    \item $S_n - f_n = \mathcal{O}\left(\frac{2^n}{n^7}\right) = o\left(\frac{2^{n+1}}{n^6}\right)$.
    \item $H_n \sim \ln n = o\left(\frac{2^{n+1}}{n^6}\right)$.
    \item $\sum R_{n,k} \le 2n 2^{n/2} = o\left(\frac{2^{n+1}}{n^6}\right)$.
\end{enumerate}
Since all residual terms are bounded by $o\left(\frac{2^{n+1}}{n^6}\right)$, we conclude that
$$ a_n = f_n + o\left(\frac{2^{n+1}}{n^6}\right) = \frac{2^{n+1}}{n} \left( 1 + \frac{1}{n} + \frac{3}{n^2} + \frac{13}{n^3} + \frac{75}{n^4} + \frac{541}{n^5} + o\left(\frac{1}{n^5}\right) \right)$$
This establishes the conjectured asymptotic expansion. \hfill $\Box$
\\\\
\textbf{Theorem.} (A conjecture of OEIS A228143, 2018) Let $s_m = \sum_{k=0}^m \binom{m}{k}^2 \binom{m+k}{k}^2$ be the sequence A005259. Let $a_n$ denote the determinant of the $(n+1) \times (n+1)$ Hankel-type matrix whose $(i, j)$-entry is $s_{i+j}$ for all $i, j = 0, \dots, n$. Let $A(x) = \sum_{n=0}^\infty a_n x^n = 1 + 48x + 161856x^2 + \dots$ denote the ordinary generating function of $a_n$. Then $A(x/3)^{1/8}$ has integer coefficients. 

\textbf{Proof.} We proceed in three main stages: analyzing the matrix entries modulo $3$, analyzing them modulo $4$, and using the resulting divisibility properties of the determinant $a_n$ to construct the required eighth root as a formal power series over the integers.

First, we determine the residues of $s_m$ modulo $3$. By writing $k$ in base $3$, Kummer's Theorem dictates that the highest power of $3$ dividing $\binom{2k}{k}$ equals the number of carries when evaluating $k + k$ in base $3$. If $k$ contains the digit $2$, a carry occurs, so $\binom{2k}{k} \equiv 0 \pmod 3$. Alternatively, if $k$ consists only of the digits $0$ and $1$, there are no carries. Applying Lucas's Theorem digit-by-digit to $\binom{2k}{k}$ yields a product of terms $\binom{0}{0}=1$ and $\binom{2}{1}=2 \equiv -1 \pmod 3$. Thus, $\binom{2k}{k} \equiv (-1)^c \pmod 3$, where $c$ is the number of $1$s in the base-3 representation of $k$. Since $c$ shares the same parity as $k$, we obtain $\binom{2k}{k} \equiv (-1)^k \pmod 3$.

We can rewrite the squared product of binomial coefficients as:
$$ \left( \binom{m}{k} \binom{m+k}{k} \right)^2 = \binom{m+k}{2k}^2 \binom{2k}{k}^2 $$
When $\binom{2k}{k} \not\equiv 0 \pmod 3$, the base-3 digits of $2k$ are strictly $0$s and $2$s. Applying Lucas's Theorem to $\binom{m+k}{2k}$, each digit of the evaluation takes the form $\binom{d_i}{0} = 1$ or $\binom{d_i}{2} \in \{0, 1\}$. Thus, $\binom{m+k}{2k} \equiv 0 \text{ or } 1 \pmod 3$. Since values in $\{0, 1\}$ are invariant under squaring, $\binom{m+k}{2k}^2 \equiv \binom{m+k}{2k} \pmod 3$. Furthermore, since $\binom{2k}{k} \in \{0, (-1)^k\} \pmod 3$, squaring it absorbs the parity: $\binom{2k}{k}^2 \equiv (-1)^k \binom{2k}{k} \pmod 3$. Combining these pieces, we find that for all $m$ and $k$:
$$ \left( \binom{m}{k} \binom{m+k}{k} \right)^2 \equiv (-1)^k \binom{m+k}{2k} \binom{2k}{k} \equiv (-1)^k \binom{m}{k} \binom{m+k}{k} \pmod 3 $$
Summing this equivalence over $k = 0, \dots, m$ yields $s_m \equiv \sum_{k=0}^m (-1)^k \binom{m}{k} \binom{m+k}{k} \pmod 3$. By a standard alternating binomial sum identity, $\sum_{k=0}^m (-1)^k \binom{m}{k} \binom{m+k}{k} = (-1)^m$. Consequently, $s_m \equiv (-1)^m \pmod 3$. 

Using this congruence, we apply row operations to the matrix $M^{(n)}$ to establish that $3^n \mid a_n$. Let $P_n$ be the lower-triangular matrix with ones on the main diagonal, $P_{i,0} = -(-1)^i$ for $i \ge 1$, and zeros elsewhere. The matrix product $P_n M^{(n)}$ replaces each row $i \ge 1$ with the $i$-th row minus $(-1)^i$ times the $0$-th row. Visually, the modified matrix takes the form:
$$ P_n M^{(n)} = \begin{bmatrix} s_0 & s_1 & s_2 & \cdots & s_n \\ s_1 + s_0 & s_2 + s_1 & s_3 + s_2 & \cdots & s_{n+1} + s_n \\ s_2 - s_0 & s_3 - s_1 & s_4 - s_2 & \cdots & s_{n+2} - s_n \\ \vdots & \vdots & \vdots & \ddots & \vdots \\ s_n - (-1)^n s_0 & s_{n+1} - (-1)^n s_1 & s_{n+2} - (-1)^n s_2 & \cdots & s_{2n} - (-1)^n s_n \end{bmatrix} $$
The new $(i, j)$-entry for $i \ge 1$ is $s_{i+j} - (-1)^i s_j$. By our previous congruence, $s_{i+j} - (-1)^i s_j \equiv (-1)^{i+j} - (-1)^i(-1)^j \equiv 0 \pmod 3$. Thus, every entry in rows $1$ through $n$ of the new matrix is divisible by $3$. Factoring a $3$ out of each of these $n$ rows gives a factor of $3^n$. Because $\det(P_n) = 1$, we conclude that $3^n \mid \det(M^{(n)}) = a_n$.

Second, we analyze $s_m$ modulo $4$. For any $k \ge 1$, the product $\binom{m}{k} \binom{m+k}{k}$ is always even, which implies its square is divisible by $4$. Thus, $\left(\binom{m}{k} \binom{m+k}{k}\right)^2 \equiv 0 \pmod 4$ for all $k \ge 1$. The $k=0$ term evaluates to $1$, yielding $s_m \equiv 1 \pmod 4$ for all $m$. 

We apply a similar matrix transformation to establish divisibility by $4^n$. Let $P'_n$ be the lower-triangular matrix with ones on the diagonal, $P'_{i,0} = -1$ for $i \ge 1$, and zeros elsewhere. The product $P'_n M^{(n)}$ subtracts the $0$-th row from every subsequent row $i \ge 1$:
$$ P'_n M^{(n)} = \begin{bmatrix} s_0 & s_1 & \cdots & s_n \\ s_1 - s_0 & s_2 - s_1 & \cdots & s_{n+1} - s_n \\ s_2 - s_0 & s_3 - s_1 & \cdots & s_{n+2} - s_n \\ \vdots & \vdots & \ddots & \vdots \end{bmatrix} $$
For $i \ge 1$, the $(i, j)$-entry becomes $s_{i+j} - s_j$. Since $s_m \equiv 1 \pmod 4$ for all $m$, this difference evaluates to $1 - 1 \equiv 0 \pmod 4$. Factoring $4$ out of the $n$ modified rows shows that $4^n \mid \det(M^{(n)}) = a_n$.

Now we combine these divisibility properties. We explicitly compute $a_1 = 48$. For $n \ge 1$, we know $4^n \mid a_n$, and since $4^n \ge 16$ for $n \ge 2$, it follows that $16 \mid a_n$ for all $n \ge 1$ (as $16 \mid 48$ for $n=1$). Because $16$ and $3^n$ are coprime, their respective divisibilities imply $16 \cdot 3^n \mid a_n$ for all $n \ge 1$.

We consider the scaled generating function $A(x/3) = \sum_{n=0}^\infty \frac{a_n}{3^n} x^n$. Let $b_n = a_n / 3^n$ denote the coefficients of this series. Since $a_0 = 1$, we have $b_0 = 1$. For $n \ge 1$, our divisibility result yields $16 \mid b_n$. Therefore, we can write $A(x/3) = 1 + 16 Y(x)$, where $Y(x) = \sum_{n=1}^\infty y_n x^n$ is a formal power series with integer coefficients and zero constant term ($Y(0) = 0$).

Finally, we construct the eighth root $C(x) = A(x/3)^{1/8} \in \mathbb{Z}\llbracket x \rrbracket$. We seek a series of the form $C(x) = 1 + 2X(x)$ where $X(x) \in \mathbb{Z}\llbracket x \rrbracket$ has no constant term. Expanding the eighth power yields
$$ \begin{aligned} (1 + 2X(x))^8 &= 1 + 16X(x) + 112X(x)^2 + 448X(x)^3 + 1120X(x)^4 \\ &\quad + 1792X(x)^5 + 1792X(x)^6 + 1024X(x)^7 + 256X(x)^8 \end{aligned} $$
Factoring out $16$ from the higher degree terms, we define the polynomial
$$ P(X) = 7X^2 + 28X^3 + 70X^4 + 112X^5 + 112X^6 + 64X^7 + 16X^8 $$
Thus, $(1 + 2X(x))^8 = 1 + 16(X(x) + P(X(x)))$. To satisfy $C(x)^8 = A(x/3)$, we must solve the equation $1 + 16(X(x) + P(X(x))) = 1 + 16Y(x)$, which simplifies to $X(x) + P(X(x)) = Y(x)$. 

We determine the coefficients $x_n$ of $X(x) = \sum_{n=1}^\infty x_n x^n$ inductively. We set $x_0 = 0$. For the inductive step, the coefficient of $x^{n+1}$ in the equation is
$$ x_{n+1} + [x^{n+1}] P(X(x)) = y_{n+1} $$
Because every term in $P(X)$ has degree at least $2$, and $X(x)$ has no constant term, the $x^{n+1}$ coefficient of $P(X(x))$ depends only on the terms $x_1, \dots, x_n$. Therefore, we can uniquely solve for the integers $x_{n+1}$ as
$$ x_{n+1} = y_{n+1} - [x^{n+1}] P\left( \sum_{k=1}^n x_k x^k \right) $$
This well-defined recurrence provides an integer sequence $(x_n)_{n \ge 1}$ giving a formal power series $X(x) \in \mathbb{Z}\llbracket x \rrbracket$ that satisfies $X(x) + P(X(x)) = Y(x)$. Substituting this back into our expansion yields $(1 + 2X(x))^8 = 1 + 16Y(x) = A(x/3)$. Setting $C(x) = 1 + 2X(x)$, we have established the existence of the desired integer power series $C(x) = A(x/3)^{1/8}$. \hfill $\Box$

These OEIS problems were added as conjectures in 2022 (\url{https://oeis.org/A051293}) and 2018 (\url{https://oeis.org/A228143}). Though neither problem has likely received much attention, the ability to provide a guarantee on the correctness of the statement that an expert conjectured serves value. For example, it appears that a good part of the work is done in identifying the asymptotic formula to be conjectured in OEIS A051293. In this case, ensuring the formula holds with a rigorous proof is a matter of validating intuition without carrying out potentially tedious proof steps, which our agent can handle. The Lean proofs for the two problems can be found at \url{https://github.com/google-deepmind/alphaproof-nexus-results/blob/main/APNOutputs/OEIS}.

\subsubsection*{A variant of the Graph Reconstruction Conjecture}
The Kelly-Ulam graph reconstruction program aims to recover a graph from its deck of one-vertex-deleted subgraphs \cite{ulam1960collection, Kelly1957}. A standard bipartite version was formulated by Bondy and Hemminger as the problem of showing that bipartite graphs are reconstructible \cite{BondyHemminger1977}. The following theorem proves a structured incidence-deletion analogue under strong type-distinguishability assumptions. 

\noindent \textbf{Terminology}
Let $\Omega=L \sqcup R$ be a finite vertex set with a partition into the disjoint parts $L,R\subseteq \Omega$. 
A graph $K$ is bipartite with respect to $(L,R)$ if every edge of $K$ has one endpoint in $L$ and one endpoint in
$R$. We say $K$ is $2$-connected if $K$ is connected and, for every vertex $v$, the induced graph on
$\Omega\setminus\{v\}$ is connected. For a graph $K$ and a vertex $v$, write $K\setminus_{\mathrm{inc}} v$
for the graph obtained from $K$ by deleting all edges incident to $v$, while
leaving the vertex set unchanged. Thus 
\[
E(K\setminus_{\mathrm{inc}} v)=\{xy\in E(K):x\ne v,\ y\ne v\}.
\]
A bipartite isomorphism $K\cong_B K'$ is a graph isomorphism whose underlying
bijection preserves both parts $L$ and $R$. The bipartite deck is the multiset
\[
\mathcal D_B(K):=\multiset{[K\setminus_{\mathrm{inc}} v]_B: v\in\Omega},
\]
where $[\cdot]_B$ denotes bipartite-isomorphism class.
Let $N_K(x)$ be the neighbor set of $x$ in $K$. The degree profile and type of
$x$ are
\[
P_K(x):=\multiset{\degG K y: y\in N_K(x)},
\qquad
\tau_K(x):=(\degG K x,P_K(x)).
\]
The type profile of $u$ is the multiset of the types of its neighbors:
\[
T_K(u):=\multiset{\tau_K(x): x\in N_K(u)}.
\]
Finally, for a type $t=(d,M)\in\mathbb N\times\operatorname{Multiset}(\mathbb N)$ and $a\in\mathbb N$, define
\[
F_a(t):=(\max(d-1,0),M\setminus\{a\}),
\]
where $M\setminus\{a\}$ means deletion of one occurrence of $a$ from the multiset
$M$.  In the applications below, $F_a$ is applied to neighbor-types, whose
first coordinate is positive, so $\max(d-1,0)=d-1$ there.  We write
$\#_t M$ for the multiplicity of $t$ in a multiset $M$.

\noindent \textbf{Theorem} (\upshape Weak bipartite graph reconstruction)
Let $G$ and $H$ be finite simple graphs on the same finite vertex set
$\Omega$, both bipartite with respect to the same parts $\Omega=L\sqcup R$ and $\card\Omega\ge 3$. Assume that $G$ is 2-connected and all vertex types of $G$ are pairwise distinct, that is, $\tau_G(x)\ne \tau_G(y)$ if  $x\ne y$. If
$\mathcal D_B(G)=\mathcal D_B(H)$,
then $G$ and $H$ are bipartite-isomorphic.

\noindent \textbf{Proof.}
Write $n=\card\Omega$ and $e(K)=\card{E(K)}$. Since bipartite isomorphisms
preserve edge counts, the deck determines the multiset
\[
\multiset{e(K\setminus_{\mathrm{inc}} v):v\in\Omega}.
\]
For every vertex $v$,
\[
e(K\setminus_{\mathrm{inc}} v)+\degG K v=e(K),                                      \tag{1}
\]
because $K\setminus_{\mathrm{inc}} v$ removes exactly the edges incident to $v$. Also,
\[
\sum_{v\in\Omega}e(K\setminus_{\mathrm{inc}} v)=(n-2)e(K),                            \tag{2}
\]
because each edge of $K$ remains in precisely the cards indexed by the
$n-2$ vertices not incident to that edge. Since $n\ge 3$, equation~(2) shows
that the deck determines $e(K)$. Applying this to the equal decks of $G$ and
$H$ gives
\[
e(G)=e(H),                                                        \tag{3}
\]
and then equation~(1) shows that the degree multisets of $G$ and $H$ are equal.

Equality of the two decks also gives a bijection $f:\Omega\to\Omega$ such that,
for every $v\in\Omega$, there is a bipartite isomorphism
\[
\phi_v:G\setminus_{\mathrm{inc}} v\cong_B H \setminus_{\mathrm{inc}}{f(v)}.                              \tag{4}
\]
For this matched pair of cards, edge counts are equal. Combining this with
(1) and (3) yields
\[
\degG G v=\degG H {f(v)}                                          \tag{5}
\]
for every $v$.

Next, $G$ has minimum degree at least $2$. Indeed, if a vertex had degree $0$,
then $G$ would not be connected; if a vertex $x$ had the unique neighbor $y$,
then the induced graph on $\Omega\setminus\{y\}$ would contain the isolated
vertex $x$, contradicting $2$-connectivity. Since the
degree multisets of $G$ and $H$ are equal, $H$ also has minimum degree at least
$2$.

Therefore, in any graph $K$ with minimum degree at least $2$, the card
$K\setminus_{\mathrm{inc}} v$ has a unique isolated vertex, namely $v$. Indeed, the vertex $v$ is isolated
because all its incident edges have been removed. On the other hand, if $x\ne v$, then $x$ has at
least two neighbors in $K$, at most one of which is $v$; hence $x$ has a
neighbor $y\ne v$, and the edge $xy$ remains in $K\setminus_{\mathrm{inc}} v$.

Applying this uniqueness to (4), the isomorphism $\phi_v$ sends the unique
isolated vertex of $G\setminus_{\mathrm{inc}} v$ to the unique isolated vertex of
$H \setminus_{\mathrm{inc}} {f(v)}$. Hence
\[
\phi_v(v)=f(v).                                                    \tag{6}
\]
Since $\phi_v$ preserves $L$ and $R$, (6) implies that $f$ preserves the two
parts:
\[
v\in L \Longleftrightarrow f(v)\in L,
\qquad
v\in R \Longleftrightarrow f(v)\in R.                              \tag{7}
\]

We now prove that $f$ preserves vertex types. It remains, in view of (5), to
compare degree profiles. For a graph $K$ and vertex $v$, put
\[
A_K(k):=\#\{x\in\Omega:\degG K x=k\},
\]
\[
B_{K,v}(k):=\#\{x\in\Omega:\deg_{K\setminus_{\mathrm{inc}} v}(x)=k\},
\]
and
\[
C_{K,v}(k):=\#\{x\in N_K(v):\degG K x=k\}.
\]
Counting vertices of degree $k$ before and after deleting the incidence set of
$v$ gives, for every $k\ge 0$,
\[
C_{K,v}(k+1)+A_K(k)+\mathbf{1}_{k=0}
=
C_{K,v}(k)+B_{K,v}(k)+\mathbf{1}_{\degG K v=k}.                           \tag{8}
\]
Indeed, in $K\setminus_{\mathrm{inc}} v$, every non-neighbor of $v$
keeps its degree, while every neighbor of $v$ has its degree lowered by
$1$.  Hence the vertices of degree $k$ in $K\setminus_{\mathrm{inc}} v$
are precisely the old non-neighbors of degree $k$, the old neighbors of
degree $k+1$, and additionally $v$ itself if $k=0$.  Therefore
\[
B_{K,v}(k)
=
A_K(k)-C_{K,v}(k)-\mathbf{1}_{\deg_K(v)=k}
+
C_{K,v}(k+1)
+
\mathbf{1}_{k=0}.
\]
Since the global
degree multisets are equal, $A_G(k)=A_H(k)$. Since $\phi_v$ is an isomorphism
between the matched cards, $B_{G,v}(k)=B_{H,f(v)}(k)$. By (5), the final
indicator in (8) is also the same for $G$ and $H$. Finally,
$C_{K,v}(0)=0$, because a neighbor always has positive degree. Induction on
$k$ in (8) gives
\[
C_{G,v}(k)=C_{H,f(v)}(k)\qquad(k\ge 0).
\]
Thus
\[
P_G(v)=P_H(f(v)),
\]
and together with (5) this proves
\[
\tau_G(v)=\tau_H(f(v))                                             \tag{9}
\]
for every $v\in\Omega$.
We next prove equality of type profiles:
\[
T_G(u)=T_H(f(u))                                                    \tag{10}
\]
for every $u$. Fix $u$. If $u\in L$, take $S=R$, otherwise take $S=L$ (thus $S$ is the
opposite side from $u$). By (7) and (9), the multisets of global types on $S$
agree:
\[
\multiset{\tau_G(x):x\in S}
=
\multiset{\tau_H(x):x\in S}.                                        \tag{11}
\]
The card isomorphism $\phi_u$ also preserves $S$, and therefore the multisets
of local types on $S$ agree:
\[
\multiset{\tau_{G \setminus_{\mathrm{inc}} u}(x):x\in S}
=
\multiset{\tau_{H \setminus_{\mathrm{inc}} {f(u)}}(x):x\in S}.                           \tag{12}
\]

For $x\in S$, bipartiteness gives the local-type formula
\[
\tau_{G \setminus_{\mathrm{inc}} u}(x)=
\begin{cases}
F_{\degG G u}(\tau_G(x)), & \text{if } x\in N_G(u),\\[1mm]
\tau_G(x), & \text{if } x\notin N_G(u).
\end{cases}                                                          \tag{13}
\]
Indeed, if $x$ is adjacent to $u$, then deleting the incidence set of $u$
removes the edge $ux$, lowering $\degG G x$ by one and deleting one occurrence
of $\degG G u$ from the degree profile of $x$. If $x$ is not adjacent to $u$,
then any neighbor of $x$ lies on the same side of the bipartition as $u$, and
therefore cannot be adjacent to $u$; hence the type of $x$ is unchanged. The
same formula holds for $H$ and $f(u)$.

Taking the number of occurrences of an arbitrary type $t$ in (13) gives
\[
\#_t\multiset{\tau_{G \setminus_{\mathrm{inc}} u}(x):x\in S}
+
\#_t T_G(u)
=
\#_t\multiset{\tau_G(x):x\in S}
+
\#_t\multiset{F_{\degG G u}(s):s\in T_G(u)}.                        \tag{14}
\]
There is an identical identity for $H$. The first terms on the two sides of
(14) agree by (12), and the global-type terms agree by (11). Also
$\degG G u=\degG H {f(u)}$ by (5). Hence equality of the type-profile counts
can be proved by descending induction on the first coordinate of $t$. The base
case is that no type in a type profile can have first coordinate at least $n$,
since every vertex degree is $<n$. For the induction step, the only types
$s$ with $F_{\degG G u}(s)=t$ have first coordinate $t_1+1$; moreover every
$s$ occurring in a type profile has positive first coordinate. Thus the
rightmost term in (14) is already equal for $G$ and $H$ by the induction
hypothesis, and subtracting the two versions of (14) yields equality of the
number of occurrences of $t$ in $T_G(u)$ and in $T_H(f(u))$. This proves (10).

The distinctness of types transfers from $G$ to $H$: if
$\tau_H(x)=\tau_H(y)$, then by applying (9) to $f^{-1}(x)$ and $f^{-1}(y)$ we
obtain equality of the corresponding $G$-types, so $x=y$.

Since types are pairwise distinct, adjacency is determined by type profiles:
for every graph $K$ among $G,H$ and every pair of vertices $a,b$,
\[
K\text{ has edge }ab
\quad\Longleftrightarrow\quad
\tau_K(b)\in T_K(a).                                                \tag{15}
\]
The forward implication is the definition of $T_K(a)$. Conversely, if
$\tau_K(b)$ occurs in $T_K(a)$, then some neighbor $c$ of $a$ has
$\tau_K(c)=\tau_K(b)$, and distinctness of types gives $c=b$.

Finally, for any $u,v\in\Omega$, equations (9), (10), and (15) give
\[
G\text{ has edge }uv
\Longleftrightarrow
\tau_G(v)\in T_G(u)
\Longleftrightarrow
\tau_H(f(v))\in T_H(f(u))
\Longleftrightarrow
H\text{ has edge }f(u)f(v).
\]
Thus $f$ is a graph isomorphism $G\cong H$. By (7), it preserves the two
bipartition parts, so it is a bipartite isomorphism. \hfill $\Box$

The Lean proof discovered for this problem is at \url{https://github.com/google-deepmind/alphaproof-nexus-results/blob/main/APNOutputs/AICollaborator/Graphs/bipartite_graph_reconstruction_conjecture_2.lean}.

\subsubsection*{Log-Concavity of Hilbert Sequences}
A pure O-sequence is the Hilbert function of a monomial Artinian level algebra; equivalently, it counts monomials by degree in a pure finite order ideal of monomials.  These sequences go back to Stanley’s work on Hilbert functions of graded algebras \cite{Stanley1978} and have since been extensively studied in combinatorial commutative algebra.  In the systematic study \cite{BMMNZ2012}, the authors proved the related unimodality theorem for pure O-sequences of codimension $3$ and type $2$.  Zanello later formulated the corresponding log-concavity conjecture and identified this same $(3,2)$ case as the main remaining open case for pure O-sequences \cite{Zanello2024}.  The following theorem proves this conjecture.

\noindent \textbf{Theorem.} (Zanello's conjecture, \cite{Zanello2024}.)
Every pure $O$-sequence of codimension $3$ and type $2$ is
log-concave.

\noindent\textbf{Terminology.}
A monomial $x_0^{m(0)}x_1^{m(1)}x_2^{m(2)}$ in three variables is
identified with its exponent vector $m=(m(0),m(1),m(2))\in\mathbb N^3$. 
We write $m'\le m$ for coordinate-wise inequality and
$|m|=m(0)+m(1)+m(2)$ for total degree.  A finite order ideal
$\Gamma\subset\mathbb N^3$ is a finite set closed downward under this
order: if $n\in\Gamma$ and $m\le n$, then $m\in\Gamma$.

A monomial $m\in\Gamma$ is maximal if every $n\in\Gamma$ with
$m\le n$ also satisfies $n\le m$.  The ideal is pure if all maximal
monomials have the same total degree; this common degree is the socle
degree.  The type is the number of maximal monomials.  The pure
$O$-sequence of $\Gamma$ is
\[
  h_\Gamma(d)=\#\{m\in\Gamma:|m|=d\}.
\]
For a monomial $g$, we simply write $h_g(d)=\#\{u\le g:|u|=d\}$. The $O$-sequence is log-concave if
\[
  h_\Gamma(i-1)h_\Gamma(i+1)\le h_\Gamma(i)^2
  \qquad(0<i<e),
\]
where $e$ is the socle degree of $\Gamma$.  We write
$\mathbf 1_P$ for the indicator of the condition $P$.

\noindent \textbf{Proof.}
Let $\Gamma$ be a pure order ideal in three variables and suppose that
its two maximal monomials are $m_1\ne m_2$.  Since $\Gamma$ is pure, $|m_1|=|m_2|=e$.
Every element of a finite order ideal lies below a maximal element, so
\[
  \Gamma=\{u:u\le m_1\} \cup \{u:u\le m_2\},
\]
After possibly interchanging $m_1$ and $m_2$, there is a coordinate
$x\in\{0,1,2\}$ such that
\[
  m_2(x)>m_1(x),
  \qquad
  m_2(y)\le m_1(y)\quad(y\ne x).                         \tag{1}
\]
Indeed, the three integers $m_1(i)-m_2(i)$ sum to zero and are not all
zero, hence one sign occurs in a minority position; if necessary we swap
the two monomials so that the minority sign is negative.

Let $c=m_1(x)+1$, and define
\[
  M_{\rm red}(y)=
  \begin{cases}
    m_2(x)-m_1(x)-1,&y=x,\\
    m_2(y),&y\ne x.
  \end{cases}
\]
Then $|M_{\rm red}|=e-c$. For $u\le m_2$, condition (1) implies
\[
  u\not\le m_1
  \quad\Longleftrightarrow\quad
  u(x)\ge c.
\]
If $d\ge c$, the map $u\mapsto u-ce_x$ bijects
$\{u\le m_2:|u|=d,\ u(x)\ge c\}$
with
$\{v\le M_{\rm red}:|v|=d-c\}$.
If $d<c$, the first set is empty.  Hence the $O$ sequence of
$\Gamma$ can be written as 
\[
  H(d):=h_\Gamma(d)
  =
  h_{m_1}(d)+\mathbf 1_{d\ge c}\,h_{M_{\rm red}}(d-c).       \tag{2}
\]
We need to prove that $H(d-1)H(d+1)\le H(d)^2$ holds for $0<d<e$.

For an integer-valued sequence $F:\mathbb N\to\mathbb Z$, define the first and scond difference sequences as
\[
  \Delta F(0)=F(0),
  \qquad
  \Delta F(d)=F(d)-F(d-1)\quad(d>0),
\]
\[
  \Delta^2F(0)=\Delta F(0),
  \qquad
  \Delta^2F(d)=\Delta F(d)-\Delta F(d-1)\quad(d>0).
\]

We first prove an elementary divisor-count formula for a single
monomial. Let $m=(\alpha,\beta,\gamma)$ be a monomial. 
For $p,q\in\mathbb N$, set
\[
  L_{p,q}(k)=\#\{r\in\mathbb N:r<p,\ r\le k,\ k-r<q\}.
\]
Thus $L_{p,q}(k)$ counts the number of pairs $(r,s)$ with
$0\le r<p$, $0\le s<q$, and $r+s=k$.  A direct box count gives
\[
  h_m(t)
  =
  \sum_{k=0}^{t}\mathbf 1_{t-k<\gamma+1}\,
  L_{\alpha+1,\beta+1}(k).
\]
With the convention $\Delta F(0)=F(0)$, the first-difference identities are
\[
  \Delta L_{p,q}(t)
  =
  \mathbf 1_{t<p}
  +
  \mathbf 1_{t<q}
  -
  \mathbf 1_{t<p+q}
  \qquad(t\ge0),
\]
and hence
\[
  \Delta h_m(t)
  =
  L_{\alpha+1,\beta+1}(t)
  -
  \mathbf 1_{t\ge \gamma+1}\,
  L_{\alpha+1,\beta+1}(t-\gamma-1)
  \qquad(t\ge 0).
\]
Taking one more difference gives, for every $t>0$,
\begin{align}
  \Delta^2 h_m(t)
  &=
  \mathbf 1_{t\le\alpha}
  +
  \mathbf 1_{t\le\beta}
  +
  \mathbf 1_{t\le\gamma}
  -
  \mathbf 1_{t\le\alpha+\beta+1}
  -
  \mathbf 1_{t\le\beta+\gamma+1}       \notag\\
  &\hspace{2.5cm}
  -
  \mathbf 1_{t\le\gamma+\alpha+1}
  +
  \mathbf 1_{t\le\alpha+\beta+\gamma+2}.              \tag{3}
\end{align}
In particular, $\Delta^2h_m(t)\le 1$. 
The same bound holds for the shifted sequence
$t\longmapsto \mathbf 1_{t\ge c}h_{M_{\rm red}}(t-c),
$
because below the shift its second difference is zero, and from the shift
onward it is the second difference of the principal sequence of
$M_{\rm red}$.  Therefore $\Delta^2H(t)\le 2$.                                      

Formula (3) also implies the following useful dichotomy.  If
$t>0$ and $\Delta^2h_m(t)=1$, then either
\[
  t\le m(0),\qquad t\le m(1),\qquad t\le m(2),              \tag{L}
\]
or
\[
  t\ge m(0)+m(1)+2,\qquad
  t\ge m(1)+m(2)+2,\qquad
  t\ge m(0)+m(2)+2,\qquad
  t\le |m|+2.                                             \tag{U}
\]
We refer to these as the lower and upper alternatives. For $t=0$, the lower alternative is automatic. 

We shall also use the following elementary consequences of the same box
count. 
\begin{enumerate}
    \item If $t\le m(i)$ for all $i$, then
\[
  2h_m(t)=(t+1)(t+2),
  \qquad
  \Delta h_m(t)=t+1.                                      \tag{5}
\]
\item If $t\ge m(i)+m(j)$ for every pair $\{i,j\}$,
and $t\le |m|+2$,
then,
\[
  2h_m(t)=(|m|+1-t)(|m|+2-t).                              \tag{6}
\]
\item If $t\ge m(i)+m(j)+1$ for every pair $\{i,j\}$,
and $t\le |m|+2$, then 
\[
  \Delta h_m(t)=-(|m|+2-t).                                \tag{7}
\]
\item For all $t$,
\[
  2h_m(t)\le (|m|-t+1)(|m|-t+2).                           \tag{8}
\]
\item If $2t\ge |m|+1$, then
\[
  \Delta h_m(t)\le0.                                      \tag{9}
\]
\end{enumerate}
Now fix $0<d<e=|m_1|=|m_2|$. Using $H(d)=H(d-1)+\Delta H(d)$
and $\Delta H(d+1)=\Delta H(d)+\Delta^2H(d+1)$
gives
\[
  H(d)^2-H(d-1)H(d+1)
  =
  (\Delta H(d))^2-H(d-1)\Delta^2H(d+1).                   \tag{10}
\]
Since $H(d-1)\ge0$ and $\Delta^2H(d+1)\le2$, it is enough to prove the following two statements:
\[
  \Delta^2H(d+1)=1
  \quad\Longrightarrow\quad
  H(d-1)\le(\Delta H(d))^2,                               \tag{11}
\]
and
\[
  \Delta^2H(d+1)=2
  \quad\Longrightarrow\quad
  2H(d-1)\le(\Delta H(d))^2.                              \tag{12}
\]
Indeed, if $\Delta^2H(d+1)\le0$, equation (10) is immediate; if it is
$1$ or $2$, then (11) or (12) applies.

Write $f(t)=h_{m_1}(t)$ and $g(t)=\mathbf 1_{t\ge c}h_{M_{\rm red}}(t-c)$, so that $H=f+g$.
Since $\Delta^2f\le1$ and $\Delta^2g\le1$, if
$\Delta^2H(d+1)=1$, then either
\[
  \Delta^2f(d+1)=1,
  \qquad
  \Delta^2g(d+1)=0,                                      \tag{A}
\]
or
\[
  \Delta^2f(d+1)=0,
  \qquad
  \Delta^2g(d+1)=1.                                      \tag{B}
\]
If $\Delta^2H(d+1)=2$, then
\[
  \Delta^2f(d+1)=\Delta^2g(d+1)=1.
  \tag{C}
\]

\medskip
\noindent\textbf{Case A1: (A)+(L) for $f$.}
Suppose equation (A) holds and the lower alternative (L) holds for $f$ with $t=d+1$. Then in particular $d+1\le m_1(x)<c$, thus $g(d-1)=g(d)=0$, hence $\Delta g(d)=0$ and therefore
\[
  H(d-1)=f(d-1),
  \qquad
  \Delta H(d)=\Delta f(d).
\]
By (5), $2f(d-1)=d(d+1)$ and $\Delta f(d)=d+1$. 
Therefore
\[
  H(d-1)=f(d-1)\le(d+1)^2=(\Delta f(d))^2=(\Delta H(d))^2.
\]
This proves (11) in this case.

\medskip
\noindent\textbf{Case A2: (A)+(U) for $f$}. Assume equation (A) holds and the upper alternative (U) holds for $f$ with $t=d+1$. Let us introduce the shorthand notation $\rho=e-d$.
Then (U) gives $\rho\ge -1$.  Applying
(6) and (7) gives
\[
  2f(d-1)=(\rho+2)(\rho+3),
  \qquad
  \Delta f(d)=-(\rho+2).                                    \tag{13}
\]
The shifted summand satisfies
\[
  \Delta g(d)\le0,
  \qquad
  2g(d-1)\le (\rho+2)(\rho+3),
  \qquad
  2g(d)\le (\rho+1)(\rho+2).        \tag{14}
\]
Indeed, the two inequalities for $g(d-1)$ and
$g(d)$ follow from the universal upper bound (8),
applied to the shifted sequence.  For the sign of $\Delta g(d)$, if
$d<c$, then $\Delta g(d)=0$.  If $d\ge c$, then the upper alternative
for $f$ implies
\[
  d\ge m_1(x)+m_1(y)+1,
  \qquad
  d\ge m_1(x)+m_1(z)+1,
\]
where $\{x,y,z\}=\{0,1,2\}$.  Adding these inequalities gives
\[
  2d\ge 2m_1(x)+m_1(y)+m_1(z)+2=e+m_1(x)+2.
\]
Since $c=m_1(x)+1$ and $|M_{\rm red}|=e-c$, this is equivalent to
$2(d-c)\ge |M_{\rm red}|+1$.  Thus the midpoint monotonicity (9) applies
to the shifted $M_{\rm red}$-sequence and gives $\Delta g(d)\le0$.

To prove (11), by (13) it remains to show
\[
  f(d-1)+g(d-1)\le\bigl(-(\rho+2)+\Delta g(d)\bigr)^2.
\]
By (13) and (14), this follows from
\[
  (\rho+2)(\rho+3)+2g(d-1)
  \le
  2\bigl(-(\rho+2)+\Delta g(d)\bigr)^2.                    \tag{15}
\]
Since $g(d)=g(d-1)+\Delta g(d)$, the third inequality in (14) gives
\[
  2g(d-1)
  \le
  (\rho+1)(\rho+2)-2\Delta g(d).
\]
Therefore the difference between the right-hand side and the left-hand
side of (15) is at least
\begin{align*}
&2\bigl(-(\rho+2)+\Delta g(d)\bigr)^2
  -(\rho+2)(\rho+3)
  -\bigl((\rho+1)(\rho+2)-2\Delta g(d)\bigr)  \\
&\qquad
=
2\Delta g(d)\bigl(\Delta g(d)-2(\rho+2)+1\bigr).
\end{align*}
This is nonnegative because $\Delta g(d)\le0$, $\rho\ge-1$, and hence
\[
  \Delta g(d)-2(\rho+2)+1\le0.
\]
Thus (11) holds in this case, and we completed the (A) case.

For the two cases when (B) holds, that is $\Delta^2g(d+1)=1$, we use the following
shift reduction.  If
\[
  \Delta^2H(d+1)=1
  \qquad\text{and}\qquad
  H(d+1)\le(\Delta H(d+1))^2,
\]
then
\[
  H(d-1)\le(\Delta H(d))^2.                                \tag{16}
\]
Indeed, $\Delta H(d)=\Delta H(d+1)-1$ and $H(d-1)=H(d+1)-\Delta H(d+1)-\Delta H(d)$.
Therefore
\[
  H(d-1)
  =
  H(d+1)-2\Delta H(d+1)+1
  \le
  (\Delta H(d+1))^2-2\Delta H(d+1)+1
  =
  (\Delta H(d))^2.
\]

\medskip
\noindent\textbf{Case B1: (B)+(L) for $g$.}
Assume equation (B) holds and the lower alternative (L) holds for $g$ with $t=d+1$. Let $D=d+1$. Since $\Delta^2g(D)=1$, necessarily $D\ge c$, and $\Delta^2h_{M_{\rm red}}(D-c)=1$.
In the lower subcase $D-c\le M_{\rm red}(i)$ for every $i$. Then $D-c+1\ge 1$, and the lower formula (5) gives
\[
  2g(D)=(D-c+1)(D-c+2),
  \qquad
  \Delta g(D)=D-c+1.                                           \tag{17}
\]
We next prove two bounds for $f$ at $D$:
\[
  f(D)\le (D+1)\Delta f(D),
  \qquad
  \Delta f(D)=c.                                       \tag{18}
\]
First, since $\Delta^2f(D)=0$, the explicit formula (3), together with
the lower inequalities for $M_{\rm red}$, implies
\[
  D\le m_1(0)+m_1(1)+1,
  \qquad
  D\le m_1(1)+m_1(2)+1,
  \qquad
  D\le m_1(0)+m_1(2)+1.                                  \tag{19}
\]
Indeed, the two pair inequalities involving $x$ follow directly from
$D-c\le M_{\rm red}(y)\le m_1(y)$ for $y\ne x$.  The remaining pair
inequality follows from (3): if it failed, then the corresponding two
single indicators would be zero; since $D\ge c=m_1(x)+1$, the
$x$-single indicator is also zero, and the right-hand side of (3)
would be negative, contradicting $\Delta^2f(D)=0$.

Thus, in (3) for $f$ at $D$, all three pair indicators and the triple
indicator are equal to $1$.  Since $\Delta^2f(D)=0$, exactly two of
the three single indicators are equal to $1$.  Moreover the
$x$-single indicator is zero, because $D\ge m_1(x)+1$.  Hence the
two remaining single indicators are equal to $1$.

It follows that for every $1\le k\le D$, the same two single
indicators are still equal to $1$, while the three pair indicators and
the triple indicator are also equal to $1$.  Therefore
\[
  \Delta^2f(k)\ge 2-3+1=0
  \qquad(1\le k\le D).
\]
Thus the first differences $\Delta f(k)$ are nondecreasing on
$\{0,1,\ldots,D\}$.  Since
\[
  f(D)=\sum_{j=0}^{D}\Delta f(j),
\]
we get
\[
  f(D)\le (D+1)\Delta f(D).
\]
This proves the first inequality in (18).

For the second inequality in (18), relabel the two coordinates different
from $x$ as $y,z$, and write
\[
  A=m_1(y)+1,
  \qquad
  B=m_1(z)+1,
\]
and recall $c=m_1(x)+1$. Using the box-count formula with $x$ as the third coordinate, the
first-difference formula gives
\[
  \Delta f(D)=L_{A,B}(D)-L_{A,B}(D-c),
\]
because $D\ge c$.  From the preceding paragraph,
$D<A$ and $D<B$, hence
\[
  L_{A,B}(D)=D+1,
  \qquad
  L_{A,B}(D-c)=D-c+1.
\]
Therefore
\[
  \Delta f(D)=(D+1)-(D-c+1)=c,
\]
proving (18). Using (17) and (18), we have
\[
  2H(D)
  =
  2f(D)+2g(D)
  \le
  2\Delta f(D)(D+1)+(D-c+1)(D-c+2).
\]
Also, by (17),
\[
  \Delta H(D)=\Delta f(D)+\Delta g(D)
  =
  \Delta f(D)+(D-c+1).
\]
Thus it is enough to show
\[
  2\Delta f(D)(D+1)+(D-c+1)(D-c+2)
  \le
  2\bigl(\Delta f(D)+D-c+1\bigr)^2.
\]
Since $D+1=(D-c+1)+c$, the difference between the right-hand side and
the left-hand side is
\[
  2\Delta f(D)\bigl(\Delta f(D)-c\bigr)
  +2\Delta f(D)(D-c+1)
  +(D-c+1)(D-c).
\]
This is nonnegative because $\Delta f(D)\ge c$ and $D-c+1\ge1$.
Therefore $H(D)\le(\Delta H(D))^2$.
By the shift reduction (16), this proves (11).

\medskip
\noindent\textbf{Case B2: (B)+(U) for $g$.}
Again let $D=d+1$ and put $\rho=e-D$.  The upper alternative for
$h_{M_{\rm red}}$ at $D-c$ gives $\rho\ge -2$.  Since
$|M_{\rm red}|=e-c$, formulas (6) and (7) give
\[
  2g(D)=(\rho+1)(\rho+2),
  \qquad
  \Delta g(D)=-(\rho+2).                                  \tag{20}
\]
For the first summand, the universal upper bound (8) gives
\[
  2f(D)\le(e-D+1)(e-D+2)=(\rho+1)(\rho+2).                 \tag{21}
\]

It remains to justify the sign of $\Delta f(D)$.  Since $g$ is in the
upper alternative at $D$, we have
\[
  D-c\ge M_{\rm red}(0)+M_{\rm red}(1)+2,\ 
  D-c\ge M_{\rm red}(1)+M_{\rm red}(2)+2,\ D-c\ge M_{\rm red}(0)+M_{\rm red}(2)+2.
\]
Adding these three inequalities gives
\[
  3(D-c)\ge 2|M_{\rm red}|+6=2(e-c)+6.
\]
Hence 
\[
  2(D-c)
  \ge
  \frac{2}{3}\bigl(2|M_{\rm red}|+6\bigr)
  =
  |M_{\rm red}|+1+\frac{|M_{\rm red}|+9}{3}
  >
  |M_{\rm red}|+1,
\]
because $|M_{\rm red}|\ge0$.
Therefore
\[
  2D=2(D-c)+2c\ge (e-c+1)+2c=e+c+1\ge e+1.
\]
By midpoint monotonicity (9), applied to $f=h_{m_1}$, we get $\Delta f(D)\le0$. Using (20), (21), we get 
\[
  2H(D)\le2(\rho+1)(\rho+2),
  \qquad
  \Delta H(D)=\Delta f(D)-(\rho+2).
\]
Since $\rho\ge-2$ and $\Delta f(D)\le0$, we have
\[
  (\rho+1)(\rho+2)\le\bigl(-\Delta f(D)+\rho+2\bigr)^2.
\]
giving the desired $H(D)\le(\Delta H(D))^2$. By the shift reduction (16), this proves (11).

\medskip
\noindent\textbf{Case C: (C) holds.}
So here $\Delta^2H(d+1)=2$ and hence $\Delta^2f(d+1)=\Delta^2g(d+1)=1$.
Since $\Delta^2g(d+1)=1$, necessarily $d+1\ge c$.
Therefore the lower alternative (L) for $f$ is impossible, because it
would imply $d+1\le m_1(x)<c$.
Hence $f$ is in its upper alternative, that is, (U) holds for $f$. The lower alternative (L) for $g$ is also impossible. Indeed, if $d+1-c\le M_{\rm red}(i)$ for every $i$,
then, choosing any $y\ne x$, we get
\[
  d+1
  \le
  m_1(x)+1+M_{\rm red}(y)
  =
  m_1(x)+1+m_2(y)
  \le
  m_1(x)+1+m_1(y).
\]
But the upper alternative for $f$ gives $d+1\ge m_1(x)+m_1(y)+2$, a contradiction.  Thus $g$ is also in its upper alternative, and (U) holds for $g$.
Put $\rho=e-d$ again. Since $|M_{\rm red}|=e-c$, we also have $|M_{\rm red}|-(d-c)=e-d=\rho$.
The upper formulas (6) and (7), applied to $f$ and to the shifted
summand $g$, give
\[
  2f(d-1)=(\rho+2)(\rho+3),
  \qquad
  \Delta f(d)=-(\rho+2),
\]
and
\[
  2g(d-1)=(\rho+2)(\rho+3),
  \qquad
  \Delta g(d)=-(\rho+2).
\]
Therefore
\[
  2H(d-1)=2(\rho+2)(\rho+3),
  \qquad
  \Delta H(d)=-2(\rho+2).
\]
Since the upper alternative for $f$ gives $\rho\ge-1$, we have
\[
  (\Delta H(d))^2-2H(d-1)
  =
  4(\rho+2)^2-2(\rho+2)(\rho+3)
  =
  2(\rho+2)(\rho+1)
  \ge0.
\]
which proves (12). This completes the proof of (11) and (12), and hence by (10) $H(d-1)H(d+1)\le H(d)^2$.
Since $0<d<e$ was arbitrary and $H$ is the pure $O$-sequence of
$\Gamma$ by (2), the theorem follows. \hfill $\Box$

The Lean proof for the problem can be found at~\url{https://github.com/google-deepmind/alphaproof-nexus-results/tree/main/APNOutputs/AICollaborator/AlgebraicGeometry}.

\subsubsection*{A Convergence Proof of Modified Anchored Gradient Descent-Ascent}
A natural-language version of the convergence proof we discovered can be found in Surina et al.~\cite{surina2026improvedlastiterateconvergencerate}.
The Lean proof is at~\url{https://github.com/google-deepmind/alphaproof-nexus-results/blob/main/APNOutputs/Optimization/LastIterateConvergence.lean}.

\subsubsection*{Written on the Wall, Conjecture 2}

For a graph $G$, let $V(G)$ be the set of vertices and $E(G)$ the set of edges. For $v \in V(G)$, write $d(v)$ for the degree of $v$, and let $N(v)$ be the set of vertices adjacent to $v$ in $G$. Let $\alpha(v)$ be the local independence number of $v$: the size of the largest independent set in $G[N(v)]$. Let $\ell(G) := \frac{1}{n}\sum_v \alpha(v)$ be the average of the local independence numbers, and let $L_S(G)$ be the maximum number of leaves in a spanning tree of $G$, or $0$ if $G$ is not connected.
\\
\textbf{Theorem.}
If $G$ is a simple connected graph on $n$ vertices, then
$$L_S(G) \geq 2(\ell(G) - 1).$$
\textbf{Proof.}
Let $V := V(G)$. For every vertex $v \in V$, fix a maximum independent set $A(v)$ in $G[N(v)]$. Then $\alpha(v) = |A(v)|$.

Now take two copies of $V$, and add edges $uv$ from $u \in V$ on the left to $v \in A(u)$ on the right. This forms a bipartite graph, where vertices on the left side have degree $\alpha(v)$. The key idea is to look at the degrees of the vertices on the right side, and formulate the entire proof in terms of them. Formally, put $c(v) := |{u \in V(G): v \in A(u)|}$ as the degree of a vertex $v$ on the right side.

Note that $c(v) \leq d(v)$, for all $v \in V$. Because $\sum_v \alpha(v) = \sum_v c(v)$, it is enough to show that $\sum_v c(v) \leq n(L_S(G)/ 2 + 1)$. To simplify the notation, we set $\mu := L_S(G)/ 2 + 1$, and so the goal is to show $\sum_v c(v) \leq n\mu$.
\\
\textbf{Lemma 1.}
For every edge $uv \in E(G)$ we have $c(u) + c(v) \leq |N(u) \cup N(v)|$.
\\
\textbf{Proof of Lemma 1.} For every vertex $w \in N(u) \cup N(v)$, at most one of $u$ and $v$ can be part of $A(w)$, otherwise the edge $uv$ would contradict the fact that $A(w)$ is independent. On the other hand, if $w \in V$ is such that $u$ or $v$ is in $A(w)$, then $w$ must be part of $N(u) \cup N(v)$. Then $c(u) + c(v)$ is at most the size of $N(u) \cup N(v)$. $\Box$
\\
\textbf{Lemma 2.}
For every edge $uv \in E(G)$ we have $|N(u) \cup N(v)| \leq L_S(G) + 2$.
\\
\textbf{Proof of Lemma 2.} We construct a spanning tree in $G$ with at least $|N(u) \cup N(v)| - 2$ leaves.

Start with the edge $uv$ and add all the vertices in $N(u) \cup N(v) \setminus \{u,v\}$ as pendant vertices of degree $1$. There are $|N(u) \cup N(v)| - 2$ leaves in this tree $T'$. Because $G$ is connected, we can expand $T'$ to a spanning tree $T$ of $G$ by adding edges until no more can be added. Then from every leaf in $T'$ there is a path to a leaf in $T$ that doesn't cross $uv$. All these leaves must be distinct, since they are contained in distinct subtrees of $T$. Then $T$ has at least $|N(u) \cup N(v)| - 2$ leaves. $\Box$
\\
Lemmas 1 and 2 show that for every edge $uv \in E(G)$,
$$c(u) + c(v) \leq L_S(G) + 2 = 2\mu.$$

Next, we observe that $\sum_v c(v) \leq n\mu$ trivially holds if every $c(v) \leq \mu$. So assume this is not the case. Then we can split the vertices into "heavy" vertices with $c(v) > \mu$, and "light" vertices with $c(v) \leq \mu$. Let $S$ be the set of heavy vertices.

$S$ must be an independent set, since the above equation shows that for any edge $uv$, at most one of $u$ or $v$ can be heavy. Less obvious is that light vertices cannot have many neighbors in $S$.
\\
\textbf{Lemma 3.} If $v$ is a light vertex, then $|N(v) \cap S| < \mu$.
\\
\textbf{Proof of Lemma 3.} Let $u \in N(v) \cap S$. Since $S$ is independent, all neighbours of $u$ are outside $N(v) \cap S$. So $|N(u) \cup N(v)| \geq |N(v) \cap S| + |N(u)|$. Using Lemma 2 we get
$$
|N(v) \cap S| \leq |N(u) \cup N(v)| - |N(u)| \leq L_S(G) + 2 - d(u).
$$
As $d(u) \geq c(u) > \mu$, we obtain $|N(v) \cap S| < L_S(G) + 2 - \mu = \mu$. $\Box$

Then the theorem follows from the following more general statement:
\\
\textbf{Claim.} Let $c: V \rightarrow \mathbb{R}$ and $\mu \geq 0$ satisfy:
\begin{enumerate}
\item $c(v) \leq d(v)$ for all $v$,
\item $c(u) + c(v) \leq 2\mu$ for every edge $uv$,
\item $S := \{u: c(u) > \mu\}$ is an independent set,
\item $|N(v) \cap S| < \mu$ for every $v \notin S$.
\end{enumerate}
Then $\sum_v c(v) \leq n\mu$.
\\
\textbf{Proof of Claim.} We assume $S$ is non-empty, otherwise the claim is trivially true. The proof uses a discharging argument: it sends the \textit{excess} $c(u) - \mu$ from a heavy vertex $u \in S$ to its neighbors. The redistribution keeps the total sum $\sum_v c(v)$ constant. However, the amount sent to a light vertex $v$ is less than the \textit{slack} $\mu - c(v)$, so the overall sum can't exceed $n\mu$.
Precisely, define a weight function that transfers excess from heavy to light vertices:
$$W(u, v) := \begin{cases}
\frac{\mu - c(v)}{d(u)} & \textrm{if $u \in S$ and $uv$ is an edge},\\
0 & \textrm{otherwise}.
\end{cases}$$
Because $S$ is independent, the weight flows only from heavy to light vertices. We now show two statements.
\\
\textbf{Each heavy vertex sends out enough weight}. If $u \in S$ and $uv$ is an edge, then $c(u) + c(v) \leq 2\mu$ by condition 2. So $\mu - c(v) \geq c(u) - \mu$. Then
$$\sum_{v \in N(u)} W(u, v) \geq d(u) \frac{c(u) - \mu}{d(u)} = c(u) - \mu.$$
\\
\textbf{Each light vertex absorbs at most its slack}. Let $v \notin S$. Then $v$ receives weight from vertices in $F := N(v) \cap S$.
\\
Let $u \in F$. From conditions (1) and (4), we have $|F| < \mu < c(u) \leq d(u)$. Then the total weight received is
$$\sum_{u \in F} W(u, v) = \sum_{u \in F} \frac{\mu - c(v)}{d(u)} < \sum_{u \in F} \frac{\mu - c(v)}{|F|} = \mu - c(v). 
$$
These two statements together show that
$$
\sum_{u \in S} c(u) - \mu \leq \sum_{u \in S} \sum_{v \notin S} W(u, v) = \sum_{v \notin S} \sum_{u \in S} W(u, v) \leq \sum_{v \notin S} \mu - c(v).
$$
Re-arranging the terms gives $\sum_v c(v) < n\mu$ as desired. \hfill $\Box$

The discovered lean proof for this is available at ~\url{https://github.com/google-deepmind/alphaproof-nexus-results/blob/main/APNOutputs/OEIS/GraphConjecture2.lean}.


\end{document}